\documentclass[]{interact}

\usepackage{epstopdf}
\usepackage[caption=false]{subfig}

\usepackage[numbers,sort&compress]{natbib}
\bibpunct[, ]{[}{]}{,}{n}{,}{,}
\renewcommand\bibfont{\fontsize{10}{12}\selectfont}
\makeatletter
\def\NAT@def@citea{\def\@citea{\NAT@separator}}
\makeatother

\theoremstyle{plain}

\theoremstyle{definition}

\theoremstyle{remark}

\usepackage{url}
\usepackage{array}
\newcolumntype{P}[1]{>{\centering\arraybackslash}p{#1}}
\newcolumntype{M}[1]{>{\centering\arraybackslash}m{#1}}
\usepackage{CJK}

\usepackage{hyperref}

\begin{document}
\title{Development of a Compliant Gripper for Safe Robot-Assisted Trouser Dressing-Undressing}

\author{
\name{Jayant Unde\thanks{This is an original manuscript of an article published by Taylor \& Francis in Advanced Robotics on 10 Jul 2024, available online: \url{https://doi.org/10.1080/01691864.2024.2376024}}, Takumi Inden, Yuki Wakayama, Jacinto Colan, Yaonan Zhu, Tadayoshi Aoyama, and Yasuhisa Hasegawa.}
\affil{Department of Micro-Nano Mechanical Science and Engineering, Nagoya University, Nagoya, Japan.}
}

\maketitle

\begin{abstract}
In recent years, many countries, including Japan, have rapidly aging populations, making the preservation of seniors’ quality of life a significant concern. For elderly people with impaired physical abilities, support for toileting is one of the most important issues. This paper details the design, development, experimental assessment, and potential application of the gripper system, with a focus on the unique requirements and obstacles involved in aiding elderly or hemiplegic individuals in dressing and undressing trousers. The gripper we propose seeks to find the right balance between compliance and grasping forces, ensuring precise manipulation while maintaining a safe and compliant interaction with the users. The gripper's integration into a custom-built robotic manipulator system provides a comprehensive solution for assisting hemiplegic individuals in their dressing and undressing tasks. Experimental evaluations and comparisons with existing studies demonstrate the gripper's ability to successfully assist in both dressing and dressing of trousers in confined spaces with a high success rate. This research contributes to the advancement of assistive robotics, empowering elderly, and physically impaired individuals to maintain their independence and improve their quality of life.
\end{abstract}

\begin{keywords}
Robotic gripper; complaint end effector; Human-robot interaction; soft and compliant robot design; robot-assisted dressing-undressing 
\end{keywords}

\section{Introduction}
The global population is experiencing a significant shift towards an aging demographic, and this trend is particularly evident in developed countries, including Japan \cite{bloom}. With advancing age, many individuals find themselves in need of assistance with their activities of daily living (ADL). In Japan alone, over 9.635 million people require support for ADL due to the challenges posed by old age and illness \cite{mofa2018}. Among the various ADLs, dressing and undressing stand out as particularly problematic for elderly individuals, especially during toileting activities. Studies conducted by Sato et al.\cite{sato2002} have revealed that individuals above the age of 70 often face difficulties in dressing themselves and frequently depend on caregivers for assistance, particularly in the context of toilet usage. Consequently, there is an urgent demand for robot-assisted trouser dressing-undressing systems to promote self-efficacy and independence among the elderly.

In recent years, robotics has emerged as a promising avenue for incorporating robotic systems in human-centric applications, extending beyond the realm of collaborative robots (co-bots) used in industrial settings to encompass social robots designed to assist patients and the elderly \cite{Broekens2009}. Despite this progress, not enough attention has been given to the critical aspect of robot end effectors for safe human-robot interaction. In scenarios where physical contact between humans and robots occurs, it is of utmost importance to ensure stable and safe interactions. The selection and design of appropriate end-effectors are, therefore, crucial to achieving safe physical human-robot interaction (pHRI).

In this article, we present the design, development, experimental evaluation, and prospective application of the gripper system, extending our previous work \cite{unde2023} by taking into account the specific needs and challenges associated with assisting hemiplegic and elderly individuals during dressing and undressing tasks. The proposed gripper aims to strike a balance between adaptability and rigidity, ensuring precise manipulation while maintaining a safe and compliant interaction with the users. By exploring and refining such gripper systems, we can advance the field of robot-assisted care for the hemiplegic or elderly and contribute to their improved quality of life.

\begin{figure}[t]
\centering
\includegraphics[scale=0.19]{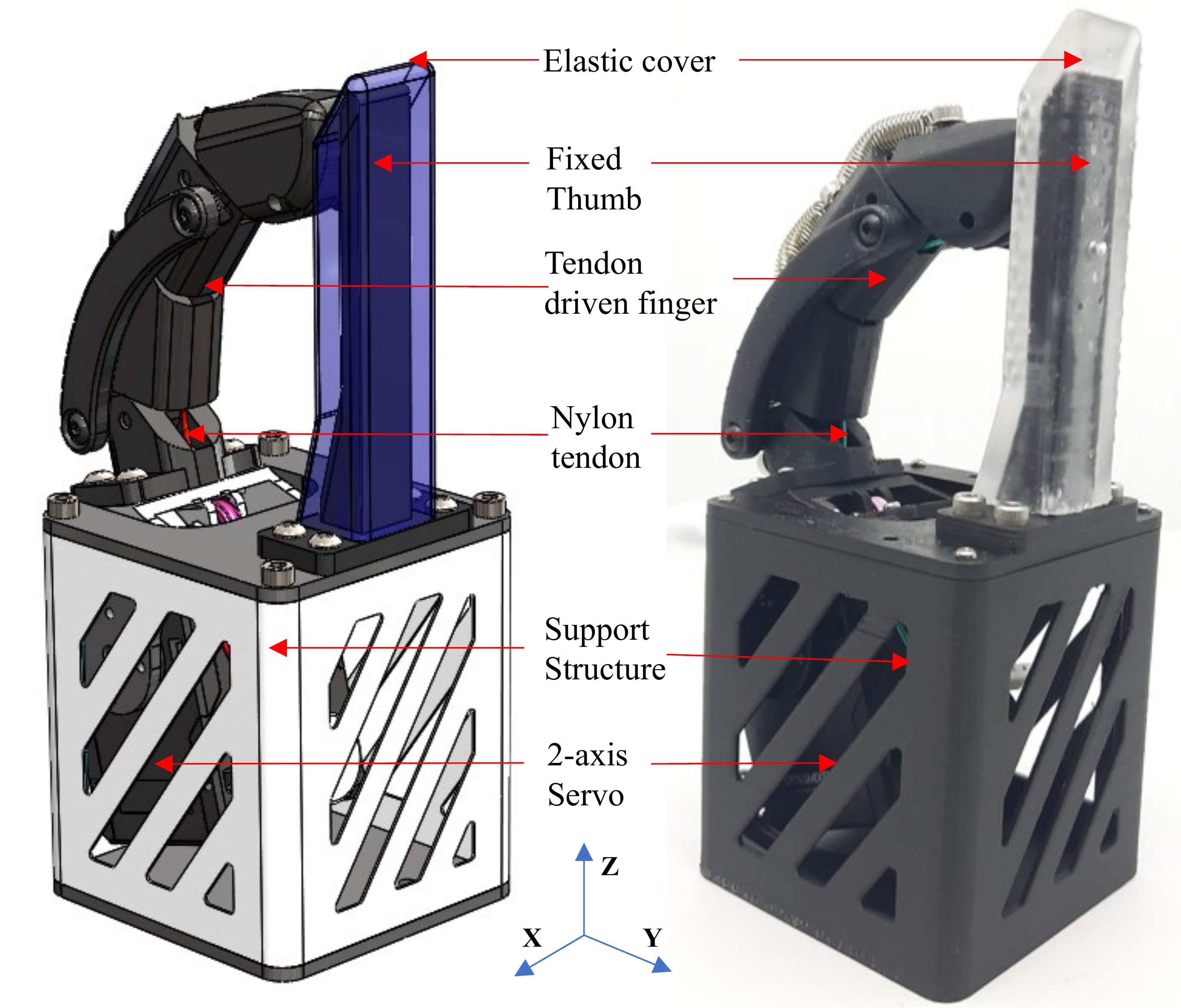}
\caption{Single actuator tendon driven linkage gripper \cite{unde2023}} \label{fig1}
\end{figure}

\textit{Contribution of the Article:}
This article makes significant contributions to the development of a specialized gripper system tailored to facilitate safe robot-assisted trouser dressing and undressing for elderly and hemiplegic individuals during toileting. The primary goal of this research is to enhance the independence and self-efficacy of elderly users during these essential activities while ensuring safe pHRI. The main contributions of this article are summarized as follows:
\begin{enumerate}
    \item Development of end effector with primary care of safe pHRI
    \item Experimental evaluation of the proposed gripper
    \item Development and demonstration of human-worn cloth manipulation for hemiplegic individual 
\end{enumerate}

\begin{table}[h]
\tbl{Overview of research studies on robotic assistance in dressing support}
{\begin{tabular}{cccc}\toprule
\textbf{Publication} & \textbf{Application} & \textbf{Modalities} & \textbf{User Test} \\ \midrule
Matsubara et al.\cite{matsubara} & Cloth estimation & Vision & No \\
Koganti et al.\cite{koganti} & Cloth state estimation & Vision & No \\
Yamazaki et al.\cite{yamazaki2013} & Cloth state estimation & Vision & Yes \\
Klee et at.\cite{klee} & Hat dressing & Vision & Yes \\
Colomé et at.\cite{colome} & Scarf dressing & Force & No \\
Tamei et at.\cite{tamei} & T-shirt dressing & Vision & No \\
Koganti et al.\cite{koganti2014} & T-shirt dressing & Vision & No \\
Chance et at.\cite{chance} & Jacket dressing & Vision, speech & No \\
Joshi et al.\cite{joshi} & Jacket Dressing & Imitation learning & Yes \\ 
Gao et al.\cite{gao} & Jacket dressing & Vision & Yes \\
Pignat et al.\cite{pignat} & Jacket and shoe dressing & Vision & No \\
Jevtić et al.\cite{jevtic} & Shoe dressing & Vision,speech & Yes \\
Yamazaki et al.\cite{yamazaki} & Trousers dressing & Vision, force & Yes \\
Hagiwara et al.\cite{hagiwara} & Trousers dressing & Vision & Yes \\ \bottomrule
\end{tabular}}
\label{table_1}
\end{table}

\section{Related Work}
Robot-assisted dressing and undressing, particularly in the context of toileting for paraplegic and elderly individuals, has been a relatively underexplored area in the existing literature. Table \ref{table_1} shows that research on clothing assistance is diverse, ranging from upper body clothing such as T-shirts and jackets to the focus of this study - trousers, and even to estimating the state of wrinkles and shape of clothing worn or on a desk.

Regarding trousers, Hagihara et al. \cite{hagiwara} developed a robot system for assisting with putting on and taking off trousers in the toilet. This study targeted people with mild paralysis who have difficulty walking. They used four robot arms to hold the trousers using the vision data and verified the process from undressing to dressing. They succeeded in 90\% of the grasping and undressing, but they also revealed the difficulty of dressing due to the elastic waistband of the trousers slipping off the gripper during the movement due to the lack of degrees of freedom (DoF) and range of motion of the two-DoF robot arm they created.

Yamazaki et al. \cite{yamazaki} also provided assistance with putting on trousers for people with mild walking difficulties using a humanoid robot. They estimated the posture of the care recipient and the state of wrinkles of the clothes and achieved a success rate of about 73\%. In this study, however, the process of dressing assistance commences with the robot already holding the clothes.

These studies indicate that research on clothing assistance is still in its infancy, but there is increasing interest in developing systems to assist individuals with disabilities. Despite this, recent studies (Table \ref{table_1}) reveal that there remains a specific lack of focus on the undressing process, particularly for elderly and paraplegic individuals. In addition, previous research has primarily focused on the development of perception and control algorithms, with less attention given to the hardware aspects of robots, especially end effectors, which are crucial for facilitating safe physical human-robot interactions.

While research on fabric manipulation grippers is abundant, there is a significant gap when it comes to the development of grippers specifically designed for human-worn cloth manipulation\cite{borras}. This unique context requires grippers with intricate capabilities to ensure safe and precise manipulation of clothing items. For instance, the gripper must possess both precision and compliance to enable safe physical human-robot interaction (pHRI). Traditional rigid two-finger grippers \cite{lundstrom1974, monkman2007} offer precise pinch grasping but lack the required compliance and adaptability for handling human-worn garments effectively \cite{birglen2018}. Conversely, compliant grippers excel at soft and adaptive cylindrical grasps, but they may not be sufficiently precise and lack a robust pinch grasp suitable for cloth manipulation tasks \cite{shintake2018}. Examples of such compliant grippers encompass tendon-driven grippers\cite{ma2015,ma2017,ko2020}, grippers with deformable structures\cite{wilson2011,belzile2013,petkovic2013,liu2017}, and soft grippers \cite{galloway2013,maccurdy2016,unde21}.

To address these challenges and fulfill the requirements for the safe manipulation of human-worn clothing, we propose a novel hybrid gripper approach that combines the benefits of both rigid and compliant grippers. By merging these two types of grippers, we aim to achieve a precise grasping capability while ensuring adequate compliance for safe human-robot interaction. Specifically, our proposed gripper utilizes a tendon-driven anthropomorphic compliant finger, which enables precise and adaptive grasping, making it well-suited for human-worn cloth manipulation during robot-assisted dressing and undressing support.

In this study, our primary focus is on the development and evaluation of this innovative gripper system, tailored to facilitate safe human-worn cloth manipulation in the context of robot-assisted dressing support. The following sections will delve into the design, methodology, and experimental evaluation of the gripper, followed by a prospective application in robot-assisted dressing-undressing during toilet usage.

\begin{figure} [t]
\centering
\subfloat[Gripper overall dimensions]{%
\resizebox*{9cm}{!}{\includegraphics{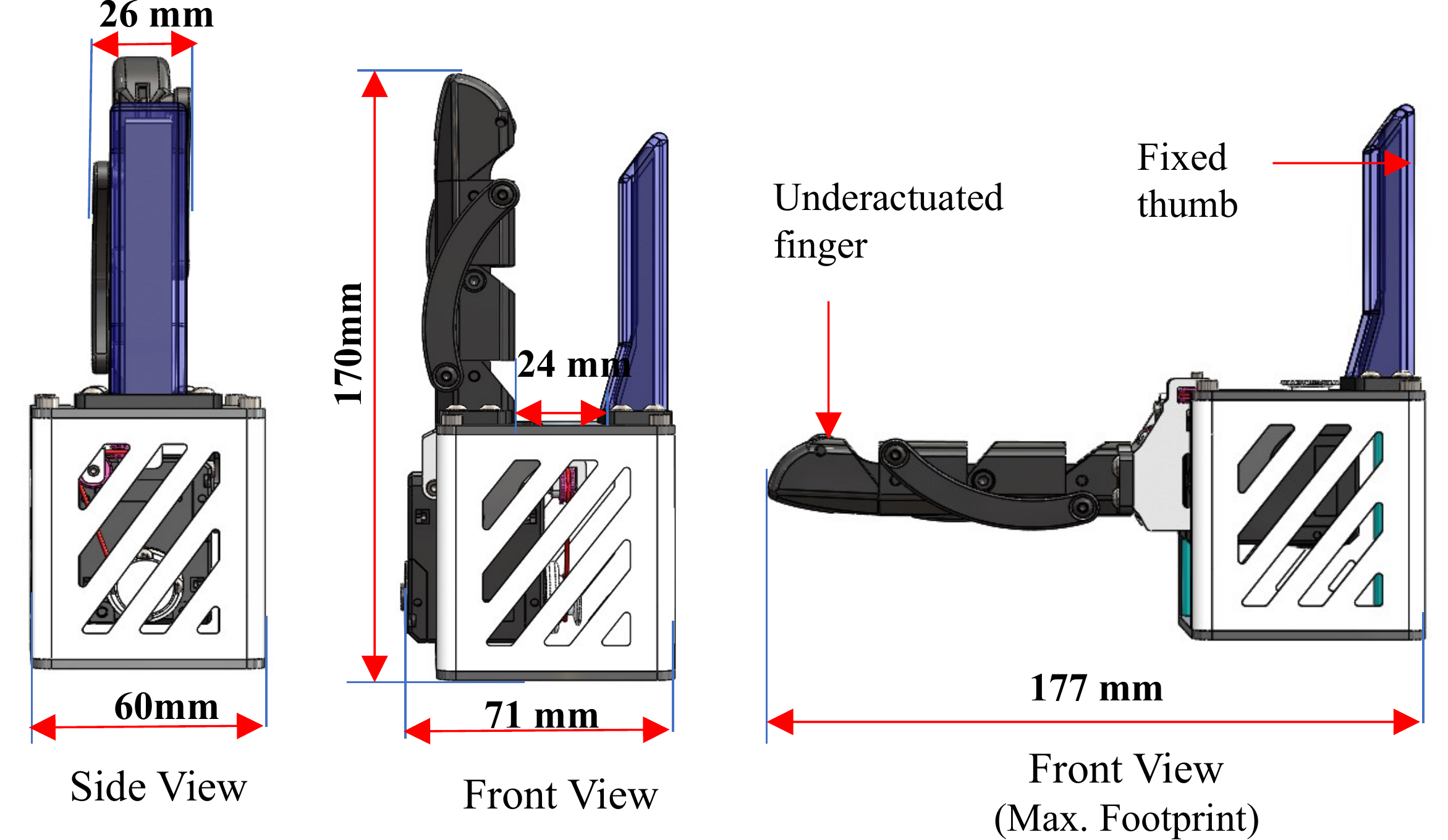}}}
\subfloat[Gripper workspace]{%
\resizebox*{5cm}{!}{\includegraphics{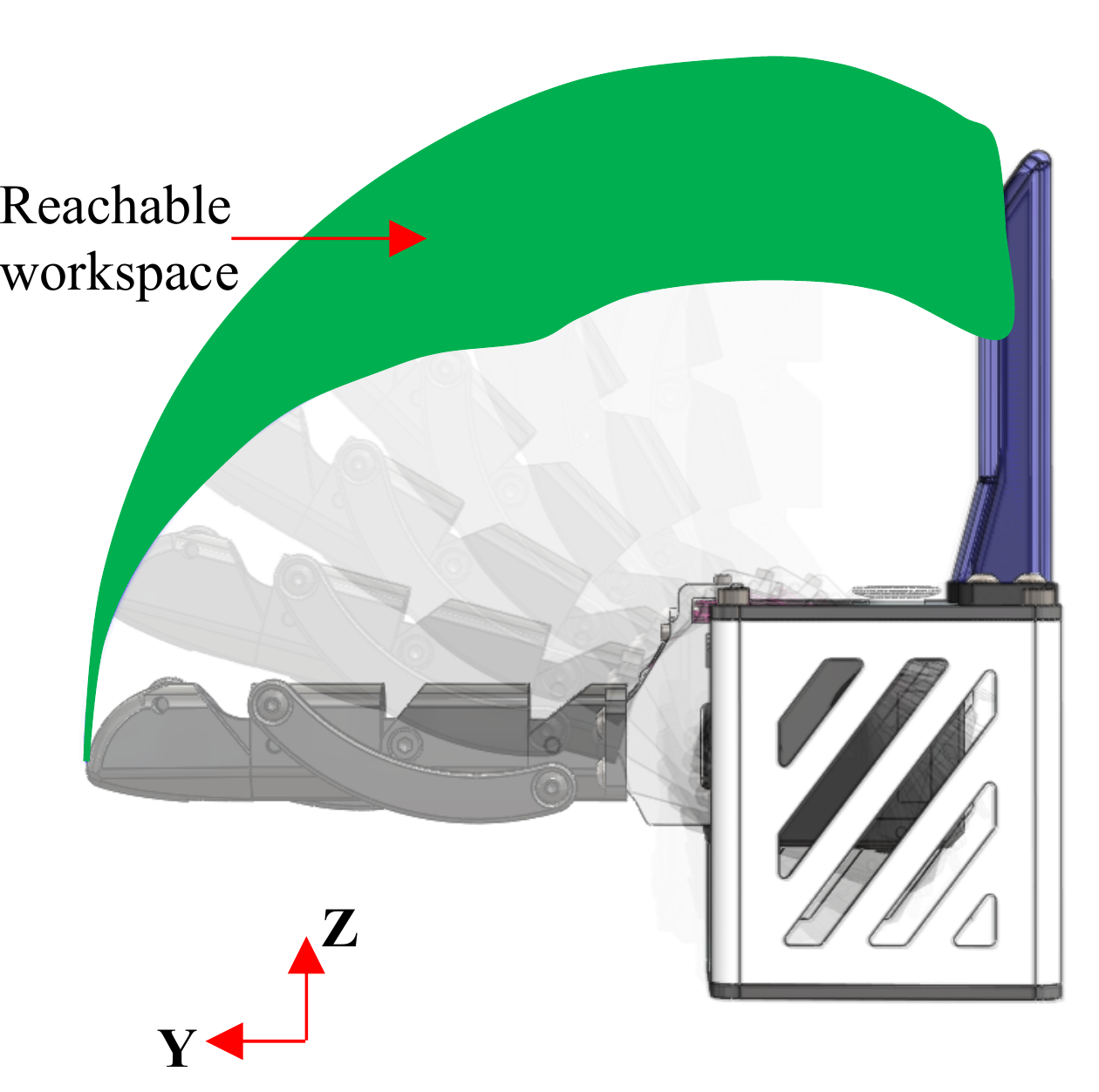}}}\hspace{3pt}
\subfloat[Finger actuation]{%
\resizebox*{7cm}{!}{\includegraphics{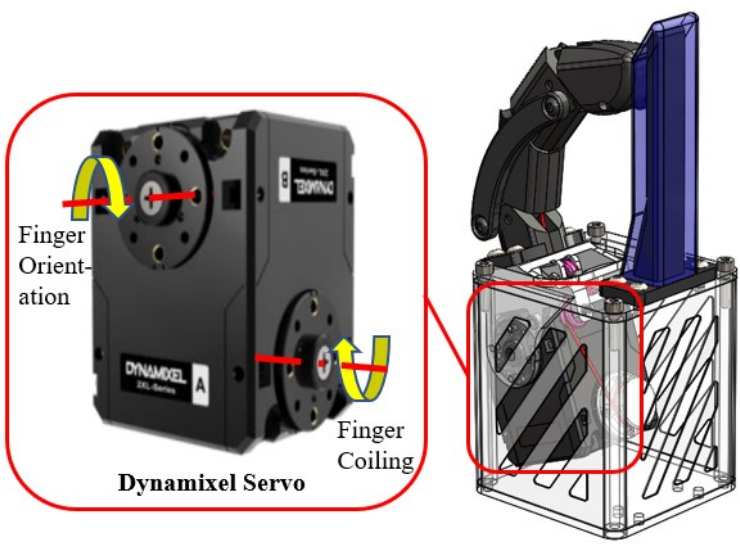}}}
\subfloat[Cylindrical grasp]{%
\resizebox*{3.5cm}{!}{\includegraphics{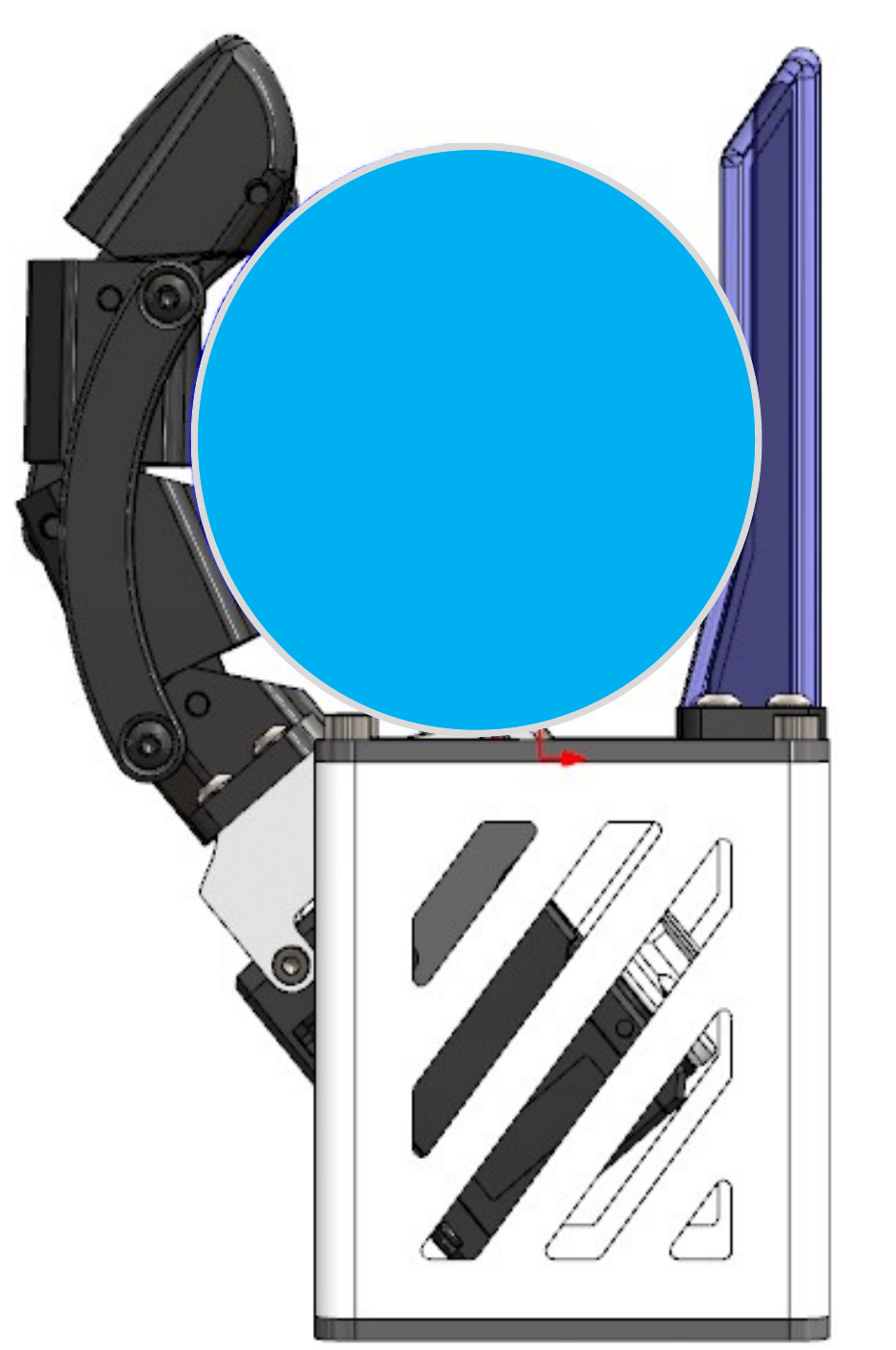}}}
\subfloat[Pinch grasp]{%
\resizebox*{2.7cm}{!}{\includegraphics{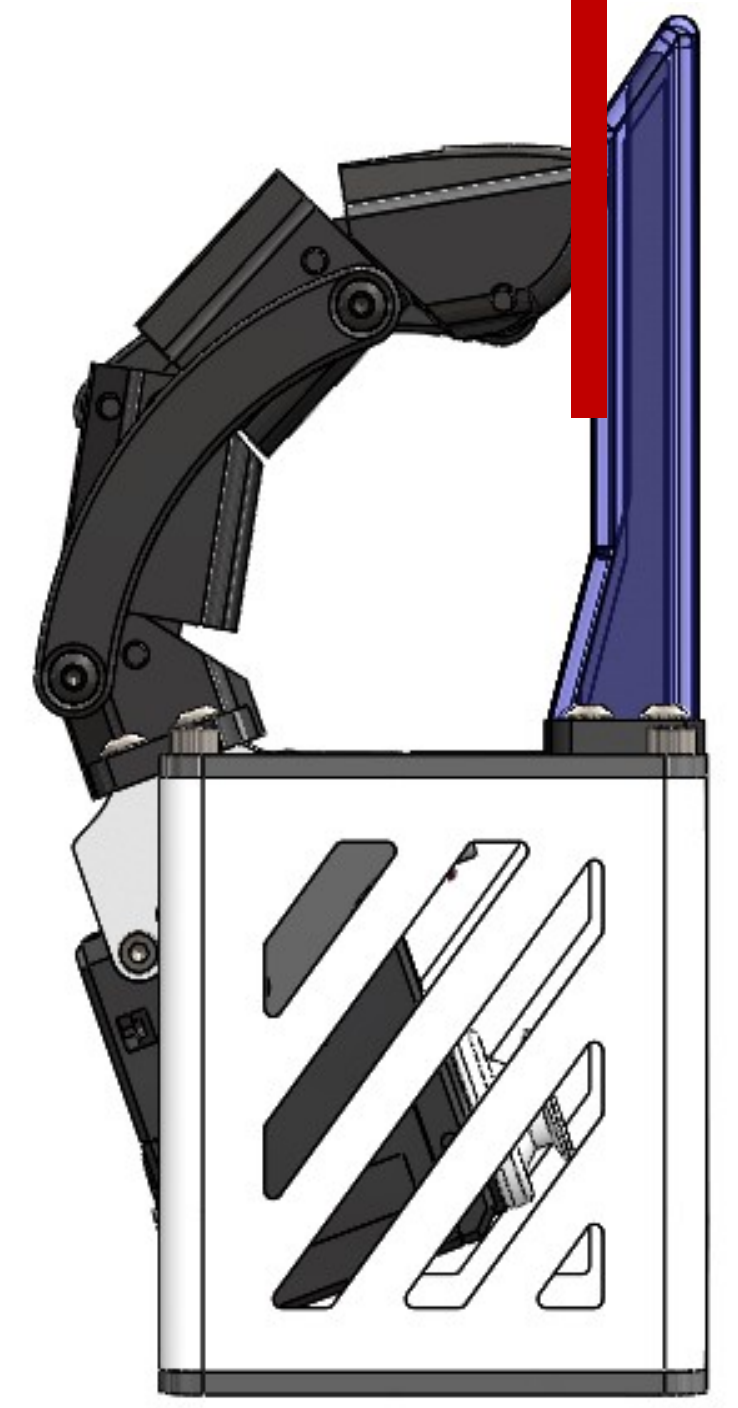}}}
\caption{Proposed gripper design} \label{fig2}
\end{figure}

 \section{Gripper Design}
 The inception of the gripper’s development was driven by the shortcomings of conventional two-finger rigid grippers and flexible robotic grippers, as discussed previously. The proposed design addresses these shortcomings by integrating rigid and flexible components to achieve both accurate pinch and adaptable cylindrical grasp in a single actuator tendon-driven two-finger linkage gripper. This distinctive characteristic allows the proposed gripper to adeptly handle trousers through a precise pinch grasp, while utilizing an adaptable cylindrical grasp for supporting limbs during the dressing process. As depicted in Figure \ref{fig1}, the gripper is composed of an anthropomorphic finger with a one DoF linkage mechanism and a fixed thumb. Figure \ref{fig2} illustrates the comprehensive workspace of the proposed gripper, demonstrating its capability to perform pinch and cylindrical grasp. The subsequent subsections will delve into these aspects in greater detail.

 \subsection{Anthropomorphic Linkage Finger Design}
The design of the gripper finger is anthropomorphic, employing two four-bar linkage mechanisms connected in series. This distinctive design limits the finger’s motion beyond a horizontal point, similar to a human finger. As shown in Figure \ref{fig2} (c), a 2-axis Dynamixel servo is utilized to operate the tendon-driven finger (Finger coiling) and to rotate the whole finger (Finger orientation). This additional DoF for finger orientation enhances the workspace and opening width of the finger.

\begin{figure}[t]
\centering
\includegraphics[scale=0.19]{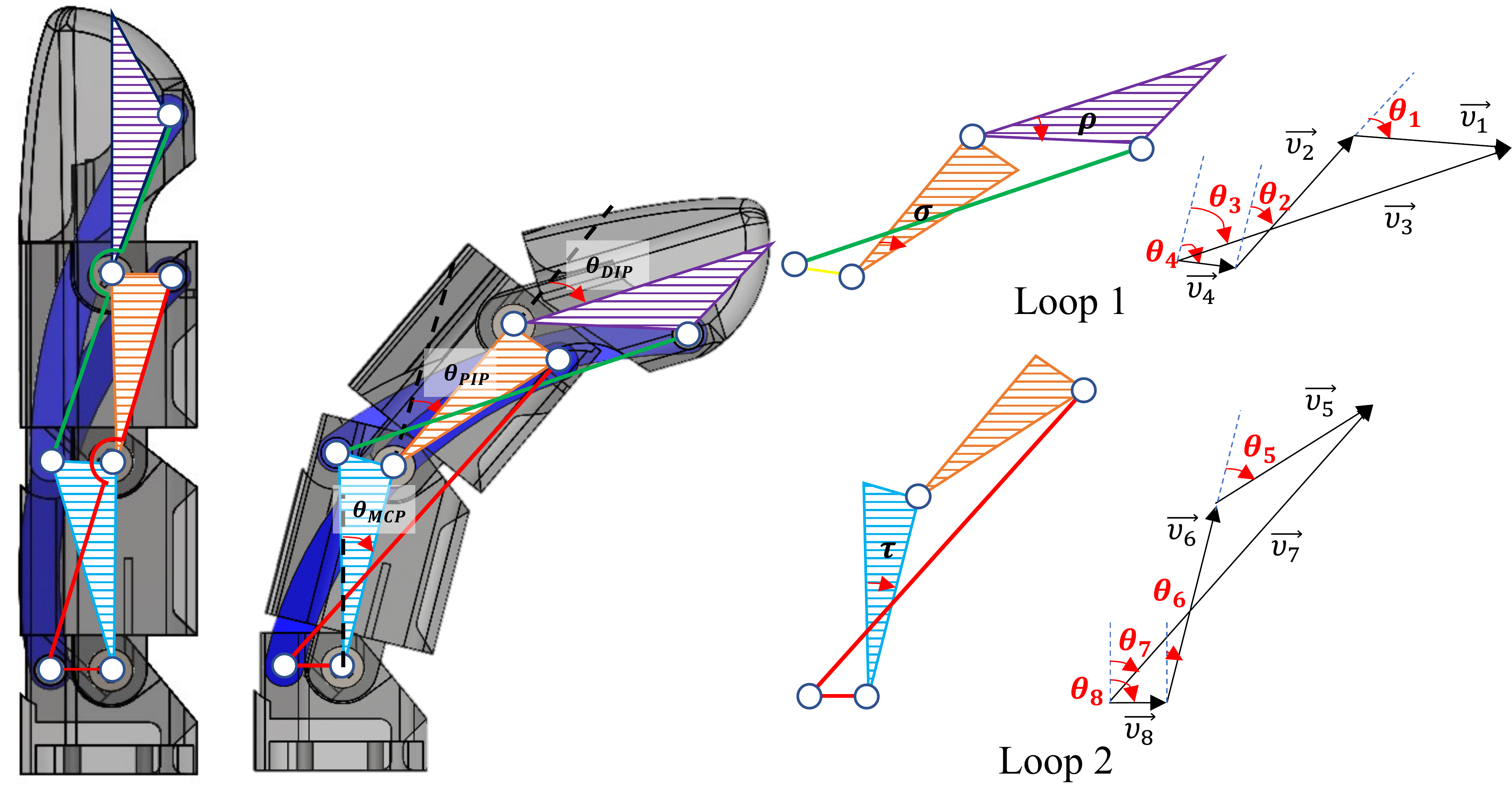}
\caption{Anthropomorphic finger four-bar linkage design and vector loops. Here, the symbols $\theta_{MCP}$, $\theta_{PIP}$, and $\theta_{DIP}$ represent the human Metacarpophalangeal (MCP) joint, Proximal Interphalangeal (PIP) joint, and Distal Interphalangeal (DIP) joint, respectively. Where,  $\theta_{MCP}=\theta_6$, $\theta_{PIP}=\theta_5-\sigma$, and $\theta_{DIP}=\theta_1-\rho$.} \label{fig3}
\end{figure}

As depicted in Figure \ref{fig3}, the linkage design emulates the motion of a human finger by mimicking the various joints of a human finger and constraining that motion through four-bar linkage mechanisms. Despite the complexity of the design, the overall degree of freedom for the linkage mechanism remains one, as indicated by the mobility equation. This suggests that the linkage mechanism can be operated by a single actuator. To drive this finger, a tendon is connected at the tip of the finger and is driven by a 2-axis Dynamixel servo motor \cite{robotis2022}.

As per Kutzbach's criterion for the mobility of a planar linkage mechanism, the mobility $M$ of the proposed linkage mechanism can be calculated as follows:
\[ M=3\left(L-1\right)-2j=3\left(6-1\right)-2\times7=1 \]
In this equation, $L$ stands for the number of links, and $j$ signifies the number of binary joints or lower pairs.

\subsubsection{Kinematic analysis of Linkage Mechanism}

In this section, we establish a mathematical correlation between the input angle of the linkage mechanism and two dependent angles. Initially, Euler's polyhedron formula \cite{mechanism2022} is used to compute the number of independent loops in a mechanism:
\[
p=j-L+1
\]
here, $p$ denotes the number of loops, $j$ represents the number of joints, and $L$ signifies the number of links.
Hence, 
$$
L=7-6+1=2
$$
This indicates that the mechanism has 2 independent loops.

As illustrated in Figure \ref{fig1}, the tendon, which is connected to the tip of the finger, will actuate the finger. Where, $ \theta_1$ is the input, while $\theta_2$ and $\theta_{6}$ are dependent on $\theta_1$.
The vector loop closure equation, depicted in Figure \ref{fig3}, is utilized to determine these relationships:
\begin{equation}
\vec{\upsilon_1}+\vec{\upsilon_2}+\vec{\upsilon_4}=\vec{\upsilon_3}    
\end{equation}
Decomposing the vector-loop closure equation into its X and Y components,
\begin{gather} 
\upsilon_1\sin{{(\theta}_1+\theta_2)}+\upsilon_2\sin{\theta_2}+\upsilon_4\sin{\theta_4}=\upsilon_3\sin{\theta_3}\\
\upsilon_1\cos{{(\theta}_1+\theta_2)}+\ \upsilon_2\cos{\theta_2}+\upsilon_4\cos{\theta_4}=\upsilon_3\cos{\theta_3}
\end{gather}
Here, substituting $\theta_4=\pi/2$ and rearranging,
\begin{gather} 
\upsilon_1\sin{{(\theta}_1+\theta_2)}+\upsilon_2\sin{\theta_2}+\upsilon_4=\upsilon_3\sin{\theta_3}\\
\upsilon_1\cos{{(\theta}_1+\theta_2)}+\ \upsilon_2\cos{\theta_2}=\upsilon_3\cos{\theta_3}
\end{gather} 
Squaring and adding equations (4) and (5),
\begin{equation} 
\begin{split}
\upsilon_3^2=\upsilon_1^2+\upsilon_2^2+\upsilon_4^2{+2\upsilon_1r}_4\sin{{(\theta}_1+\theta_2)}+{2\upsilon_2\upsilon}_4\sin{\theta_2}{+\ 2\upsilon_1\upsilon}_2\cos{\theta_1}
\end{split}
\end{equation}
Dividing by ${2\upsilon_1\upsilon}_2$ and rearranging equation (6),
\begin{equation} 
\begin{split}
\frac{\upsilon_1^2+\upsilon_2^2-\upsilon_3^2+\upsilon_4^2}{{2\upsilon_1\upsilon}_2}+\frac{\upsilon_4}{\upsilon_2}\sin{{(\theta}_1+\theta_2)}+\frac{\upsilon_4}{\upsilon_1}\sin{\theta_2}+\cos{\theta_1}=0
\end{split}
\end{equation}
Substituting the following in equation (7),
\begin{align*}
\begin{gathered}
\kappa_1=\frac{\upsilon_4}{\upsilon_2};\kappa_2=\frac{\upsilon_4}{\upsilon_1}; \kappa_3=\frac{\upsilon_1^2+\upsilon_2^2-\upsilon_3^2+\upsilon_4^2}{{2\upsilon_1\upsilon}_2}
\end{gathered}    
\end{align*}

\begin{equation} 
\begin{split}{\kappa_3+\kappa}_1\sin{{(\theta}_1+\theta_2)}+\kappa_2\sin{\theta_2}+\cos{\theta_1}=0
\end{split}
\end{equation}

To solve equation (8) for $\theta_2$ in terms of $\theta_1$,  We utilize the following trigonometric identities:
\begin{align*}
\sin{\theta_2}=\frac{2\tan{\frac{\theta_2}{2}}}{1+\tan^2{\frac{\theta_2}{2}}};\ \cos{\theta_2}=\frac{1-\tan^2{\frac{\theta_2}{2}}}{1+\tan^2{\frac{\theta_2}{2}}}
\end{align*}
After substitution and rearranging, equation (8) becomes,
\begin{equation} 
\begin{split}
\left[\cos{\theta_1-\kappa_1\sin{\theta_1}}+\kappa_3\right]\tan^2{\frac{\theta_2}{2}}\\+[2\ \kappa_1\cos{\theta_1}+2\ \kappa_2] \tan{\frac{\theta_2}{2}}\ \\+ [\cos{\theta_1}+\kappa_1\sin{\theta_1}+\kappa_3]=0
\end{split}
\end{equation}
Substituting the following in equation (9),
\begin{align*}
\alpha=\cos{\theta_1-\kappa_1\sin{\theta_1}}+\kappa_3;\\
\beta=2 \kappa_1\cos{\theta_1}+2\ \kappa_2; \\
\gamma=\cos{\theta_1}+\kappa_1\sin{\theta_1}+\kappa_3       
\end{align*}
\begin{equation} 
\alpha\tan^2{\frac{\theta_2}{2}}+\beta\tan{\frac{\theta_2}{2}}+\gamma=0    
\end{equation}
Equation (10) is a quadratic equation, hence $\theta_2$ is,
\begin{equation}
\theta_2=2\tan^{-1}{\left(\frac{- \beta+\sqrt{\beta^2-4\alpha \gamma}}{2\alpha}\right)}    
\end{equation}

For loop 2, the vector-loop closure equation is as follows, 

\begin{equation}
\vec{\upsilon_5}+\vec{\upsilon_6}+\vec{\upsilon_8}=\vec{\upsilon_7}\ \ 
\end{equation}

We can solve the above loop closure equation similar to loop 1, where $\theta_6$ is unknown and $\theta_5=\theta_2+\sigma$ is an input angle;
\begin{gather}
\theta_6 = 2\tan^{-1}\left(\frac{-\epsilon + \sqrt{\epsilon^2-4\delta \zeta}}{2\delta}\right)    
\end{gather}

where,
\begin{align*}
\delta=\cos{\theta_5-\kappa_4\sin{\theta_5}}+\kappa_6;\\
\epsilon=2\ \kappa_4\cos{\theta_5}+2\ \kappa_5;\\
\zeta=\cos{\theta_5}+\kappa_4\sin{\theta_5}+\kappa_6
\end{align*}
and,
\begin{align*}
\begin{gathered}
\kappa_4=\frac{\upsilon_8}{\upsilon_6};\kappa_5=\frac{\upsilon_8}{\upsilon_5}{;\ \kappa}_6=\frac{\upsilon_5^2+\upsilon_6^2-\upsilon_7^2+\upsilon_8^2}{{2\upsilon_5\upsilon}_6}
\end{gathered}
\end{align*}

\begin{figure}[t]
\centering
\subfloat[Joint angles: model vs CAD simulation]{%
\resizebox*{6.6cm}{!}{\includegraphics{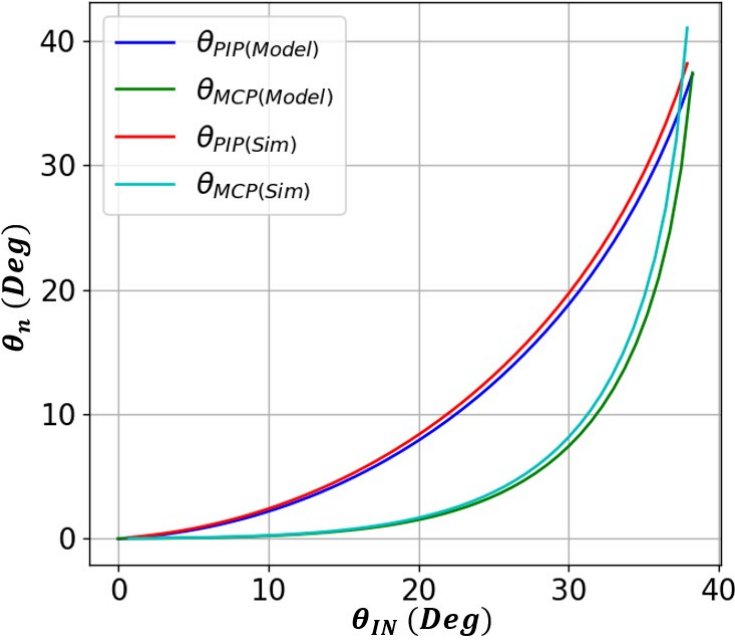}}}
\subfloat[Tip traces of fingertip: model vs CAD simulation]{%
\resizebox*{7cm}{!}{\includegraphics{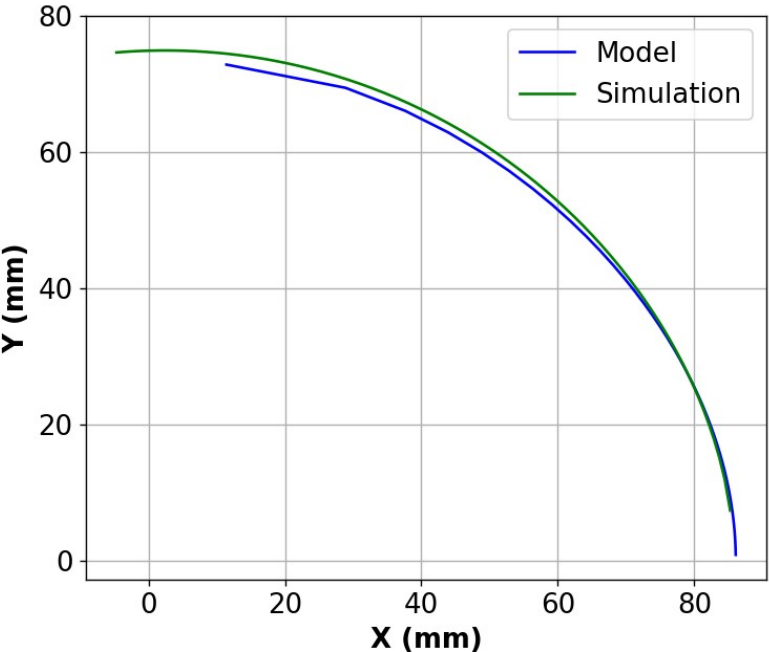}}}
\caption{Result of mathematical model and motion analysis simulation} \label{fig4}
\end{figure}

The kinematic analysis of the linkage mechanism reveals that, 
$$
\theta_2 = f(\theta_1) \quad \text{and} \quad \theta_6 = f(\theta_2).
$$
In addition, we carried out a motion analysis of the proposed anthropomorphic finger using SolidWorks. As depicted in Figure \ref{fig4} (a) \& (b), there is a significant correlation between the analytical model results and the simulation outcomes.

\subsection{Types of Actuation}
The actuation of the finger is facilitated by a tendon attached at its tip. Based on whether the closing and opening motion of the finger is passively or actively actuated, we have designed the following two types of actuation mechanisms:
\begin{enumerate}
  \item Single Tendon Driven Actuation
  \item Double Tendon Driven Actuation
\end{enumerate}

\begin{table}[b]
\centering
\tbl{Comparison of single and double tendon actuation grippers}
{\begin{tabular}{ccc}
\toprule
\textbf{Features} & \textbf{Single Tendon} & \textbf{Double Tendon}\\
\midrule
 & \includegraphics[scale=0.2]{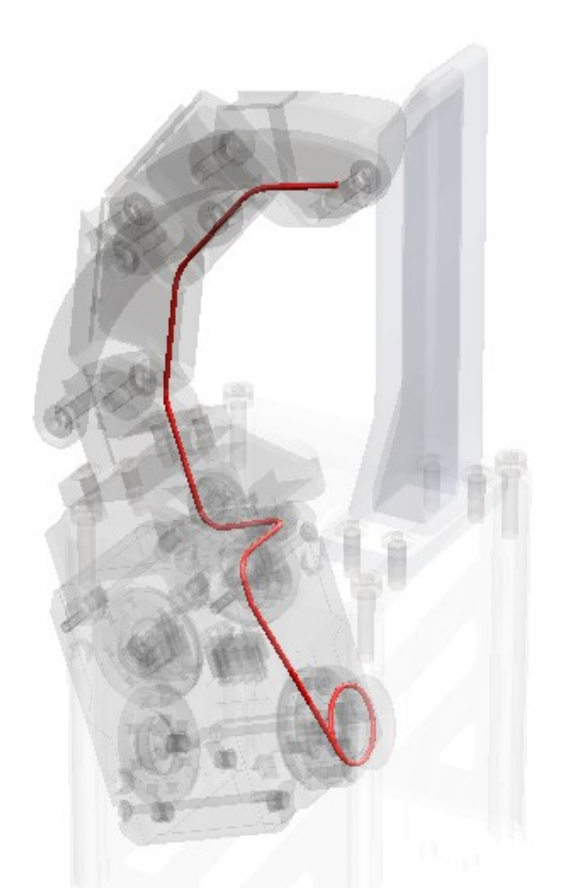} & \includegraphics[scale=0.2]{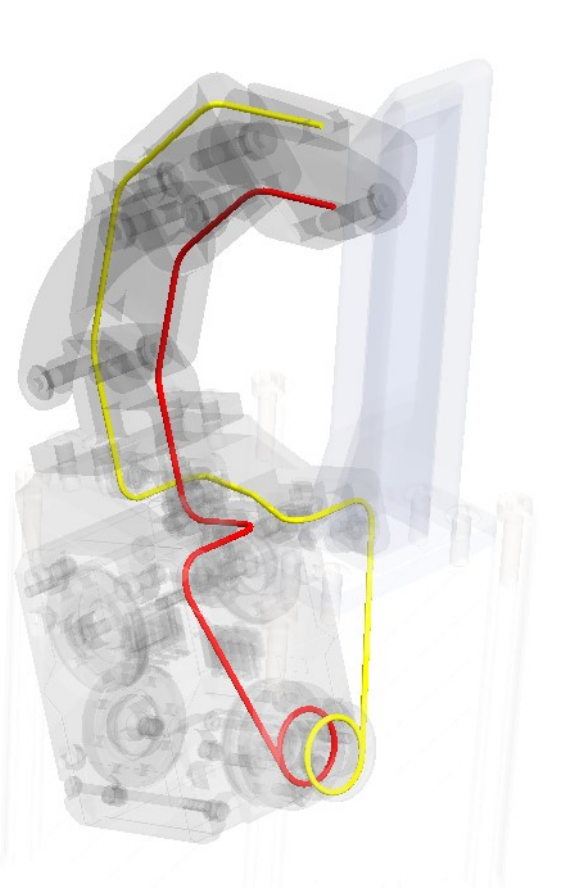}\\
Compliance & Embodied & Medium\\
Actuation Direction Control & Single (Closing) & Dual (Both Opening \& Closing)\\
Pinch Force & 7.8 N & 11.8 N\\
Actuation Force & High (Spring force) & Low (No Springs)\\
External Force Compliance & High & Medium\\
Built-in Compliance & High & Low\\
Suitability for Application & Safe pHRI & Precise control of workspace\\
\bottomrule
\end{tabular}}
\label{table_2}
\end{table}

\begin{figure}[t]
\centering
\subfloat[Single tendon driven]{%
\resizebox*{7cm}{!}{\includegraphics{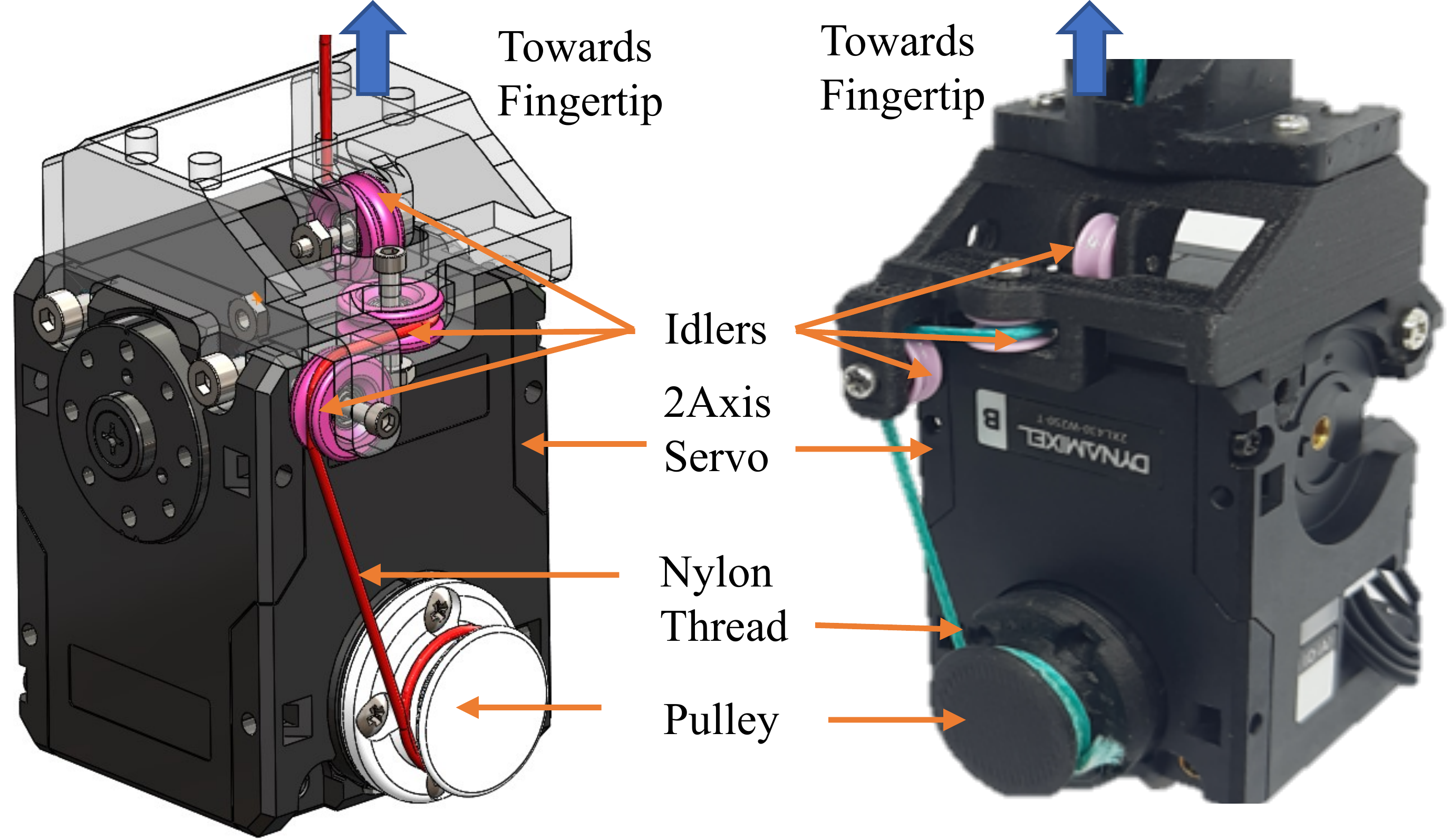}}}
\subfloat[Double tendon driven]{%
\resizebox*{7cm}{!}{\includegraphics{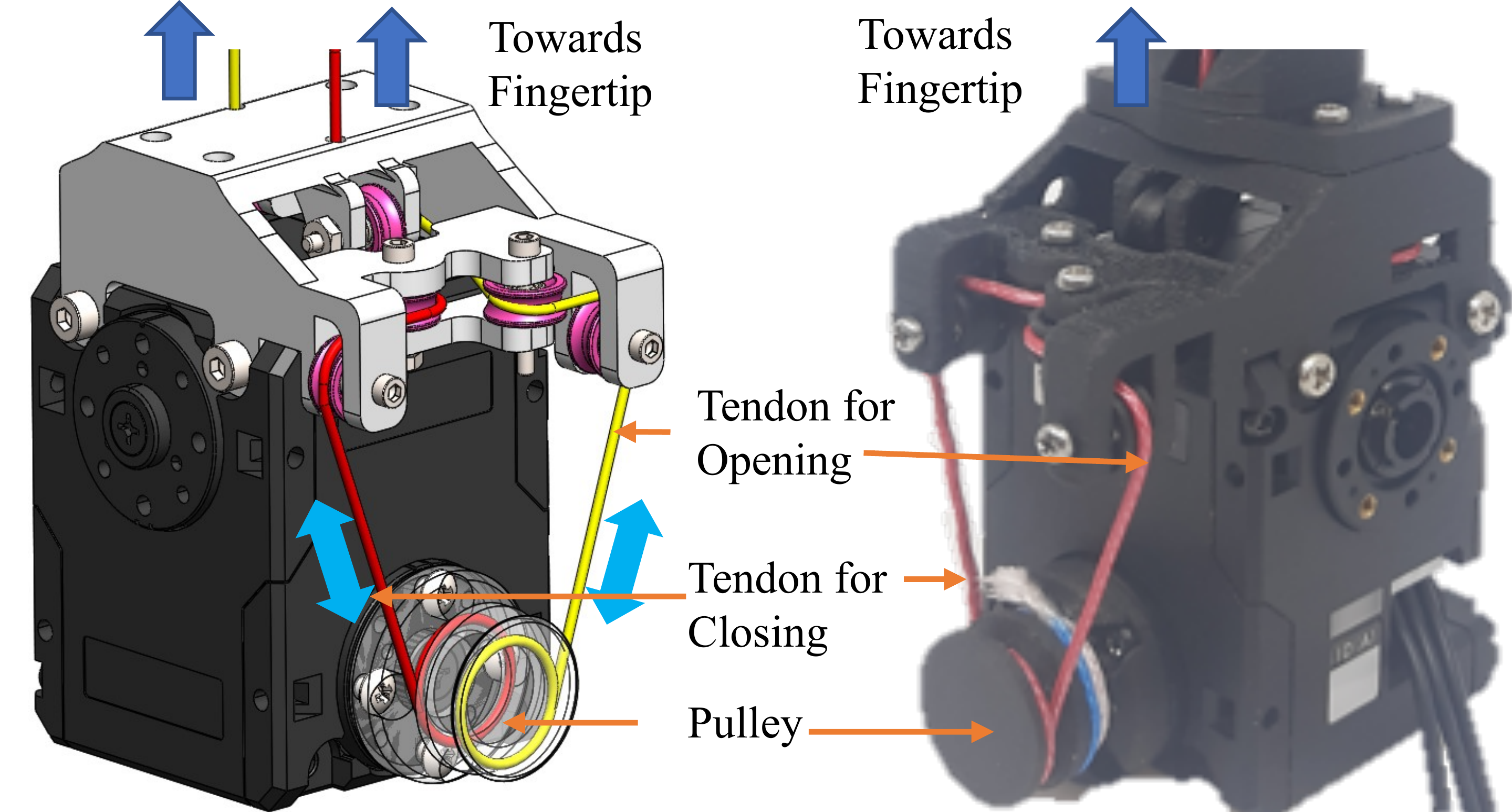}}}
\caption{Types of finger actuation mechanisms} \label{fig5}
\end{figure}

In a single tendon-driven finger shown in Figure \ref{fig5} (a), the closing motion is achieved through the tendon attached at the finger's tip. In contrast, the opening motion is passively facilitated by extension springs attached to the finger's back. This design is advantageous as it is compliant with external forces, making it suitable for close human contact and safe-pHRI.

On the other hand, a double tendon-driven finger achieves both opening and closing motions through tendon tension. The arrangement of the tendon for this actuation is depicted in Figure \ref{fig5} (b). This design allows for precise control within the finger's workspace but lacks the feature of compliance to external force found in a single tendon-driven finger. Nonetheless, the use of strings introduces a certain degree of inherent compliance. A comparison between single tendon-driven actuation and double tendon-driven actuation is presented in Table \ref{table_2}.

\subsection{Fixed Thumb Design}
The fixed thumb, devoid of actuation, is engineered to execute both pinch and cylindrical grasps. It is fabricated using 3D printing technology with Thermoplastic Polyurethane (TPU), a material recognized for its superior biocompatibility and flexibility. In addition, the gripper thumb is enveloped in a soft silicone material, enhancing safe physical human-robot interaction and providing the gripper with increased adaptability. Furthermore, it is modular in design, allowing for quick replacement based on the application requirements.

\begin{figure}[t]
\centering
\subfloat[Gripper overall footprint]{%
\resizebox*{10cm}{!}{\includegraphics{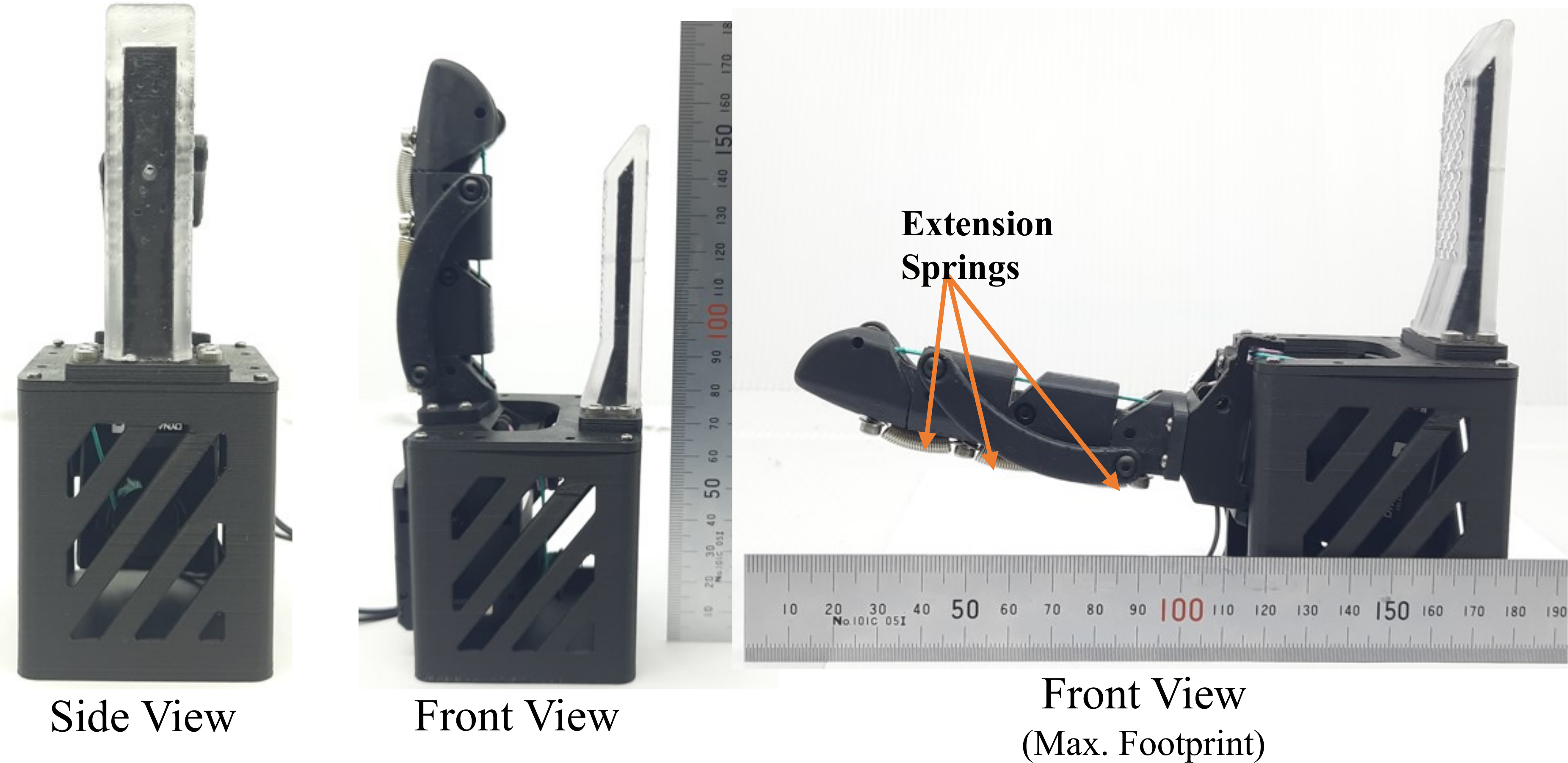}}}\hspace{2cm}
\subfloat[Compliance to external force]{%
\resizebox*{4.5cm}{!}{\includegraphics{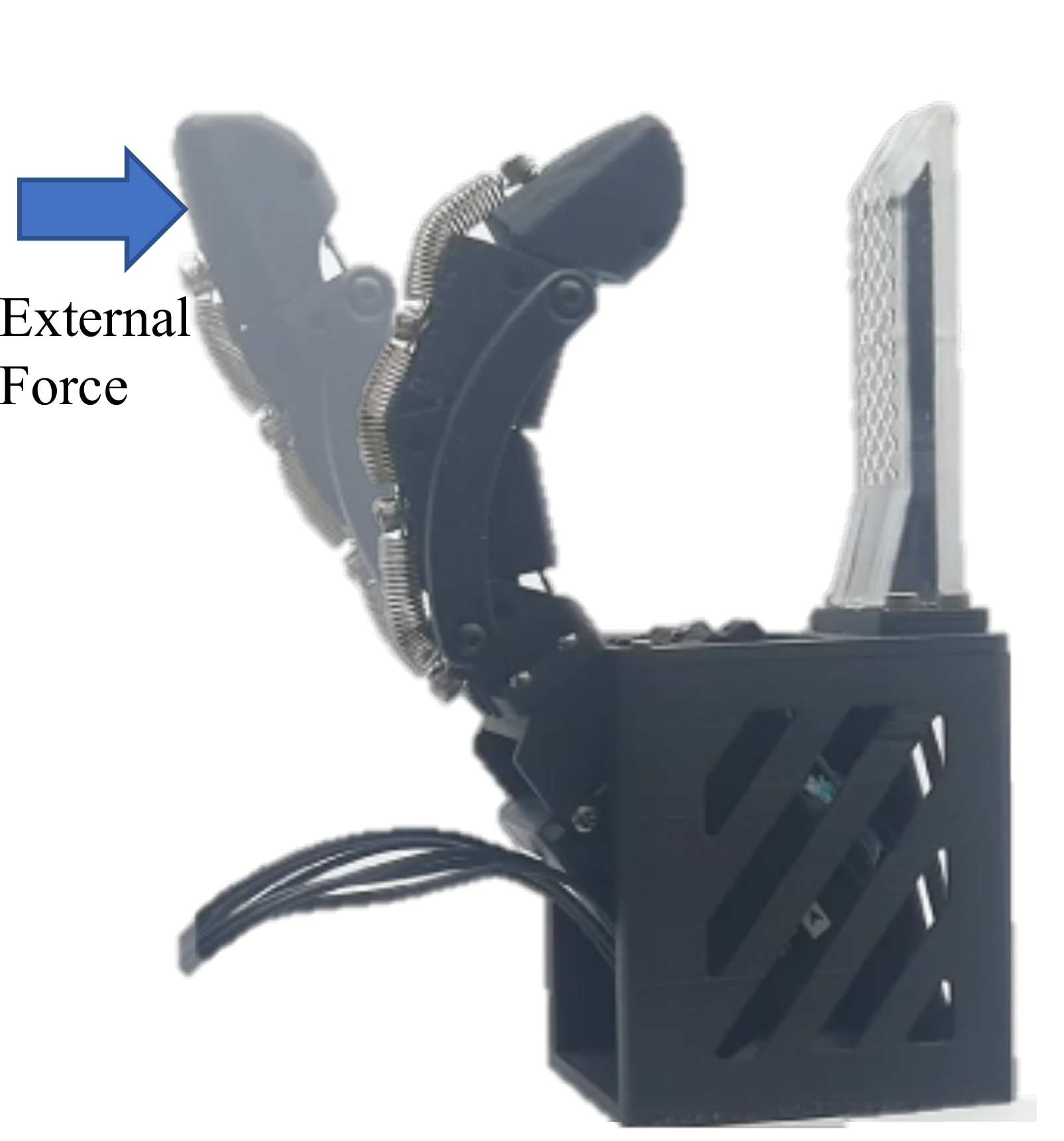}}} 
\subfloat[Cylindrical grasp]{%
\resizebox*{3cm}{!}{\includegraphics{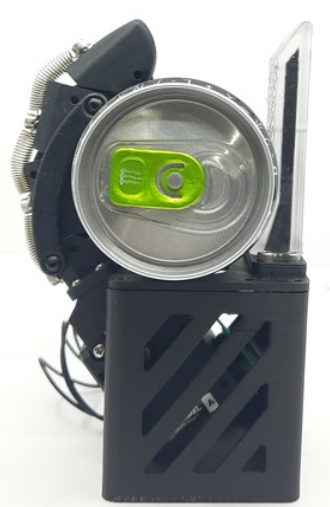}}}
\subfloat[Pinch grasp]{%
\resizebox*{2.7cm}{!}{\includegraphics{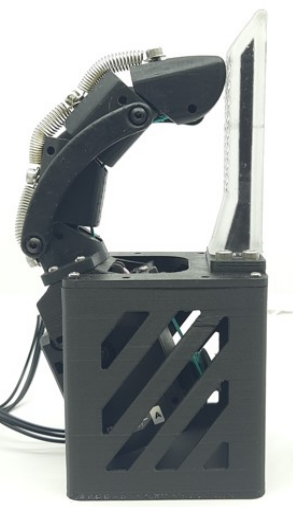}}}
\caption{Prototype of proposed gripper} \label{fig6}
\end{figure}

\subsection{Gripper Prototype}
As depicted in Figure \ref{fig6}, the prototype of the proposed gripper is manufactured using 3D printing, facilitating swift prototyping. The gripper's support structure is 3D printed with Onyx \cite{onyx}, a nylon composite material known for its robust structural strength. In contrast, the tendon-driven finger is 3D printed using the Stereolithography (SLA) printing method with Formlabs BioMed black resin material \cite{biomed}. This material is biocompatible, making it suitable for applications necessitating long-term skin contact. As previously stated, the stationary thumb is 3D printed with TPU, an elastic material that ensures safe and compliant human interaction. The total weight of the prototype, inclusive of the servo motor for actuation, is around 235 grams.

\section{Experimental Evaluation of Proposed Gripper}
To assess the effectiveness of the proposed gripper, we carried out a series of experiments.

\begin{figure}[t]
\centering
\begin{minipage}{.5\textwidth}
\flushright
\subfloat[Experimental setup]{%
\resizebox*{8cm}{!}{\includegraphics{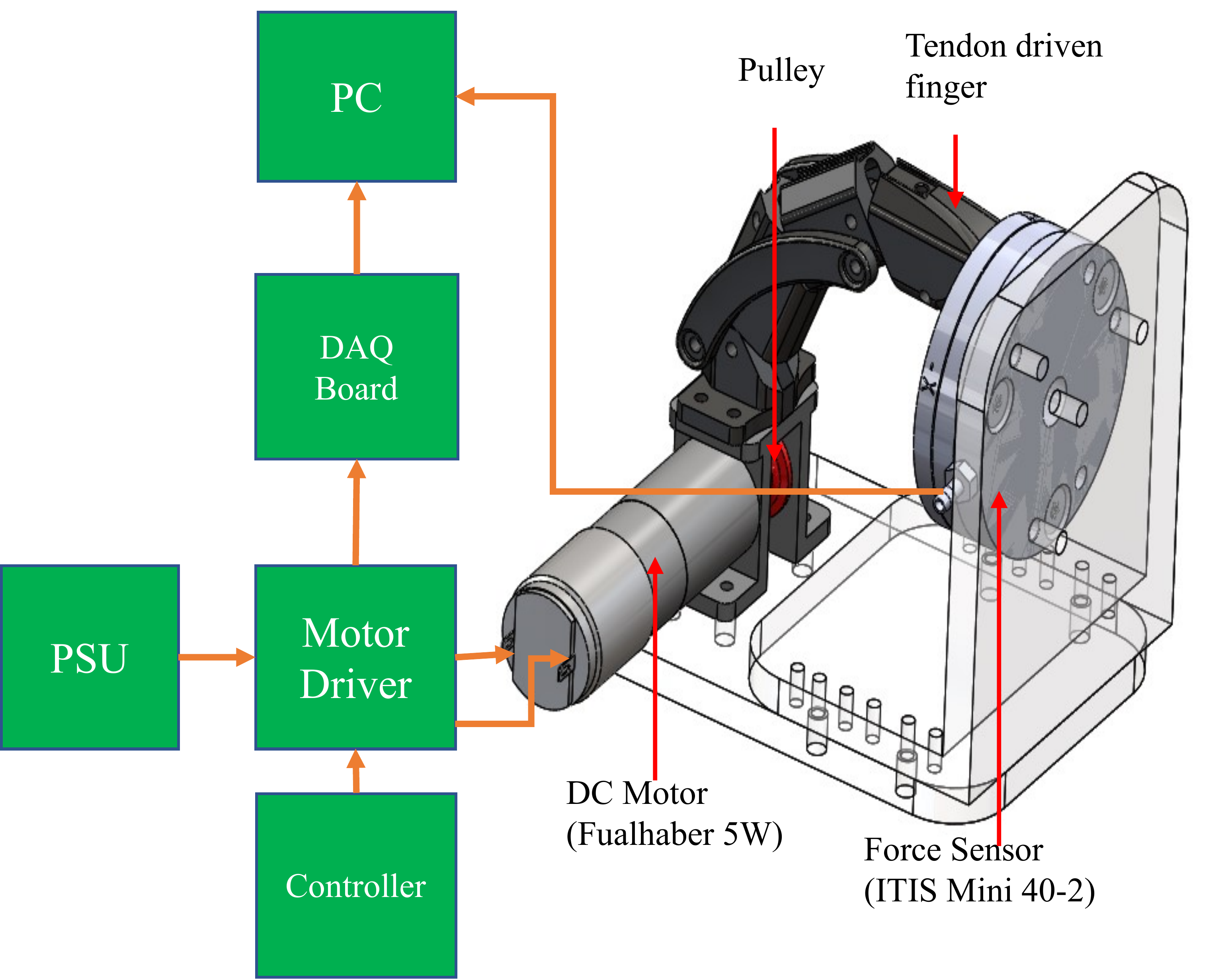}}}\hspace{5cm}
\end{minipage}%
\begin{minipage}{.42\textwidth}
\flushright
\subfloat[Proposed linkage finger]{%
\resizebox*{5.2cm}{!}{\includegraphics{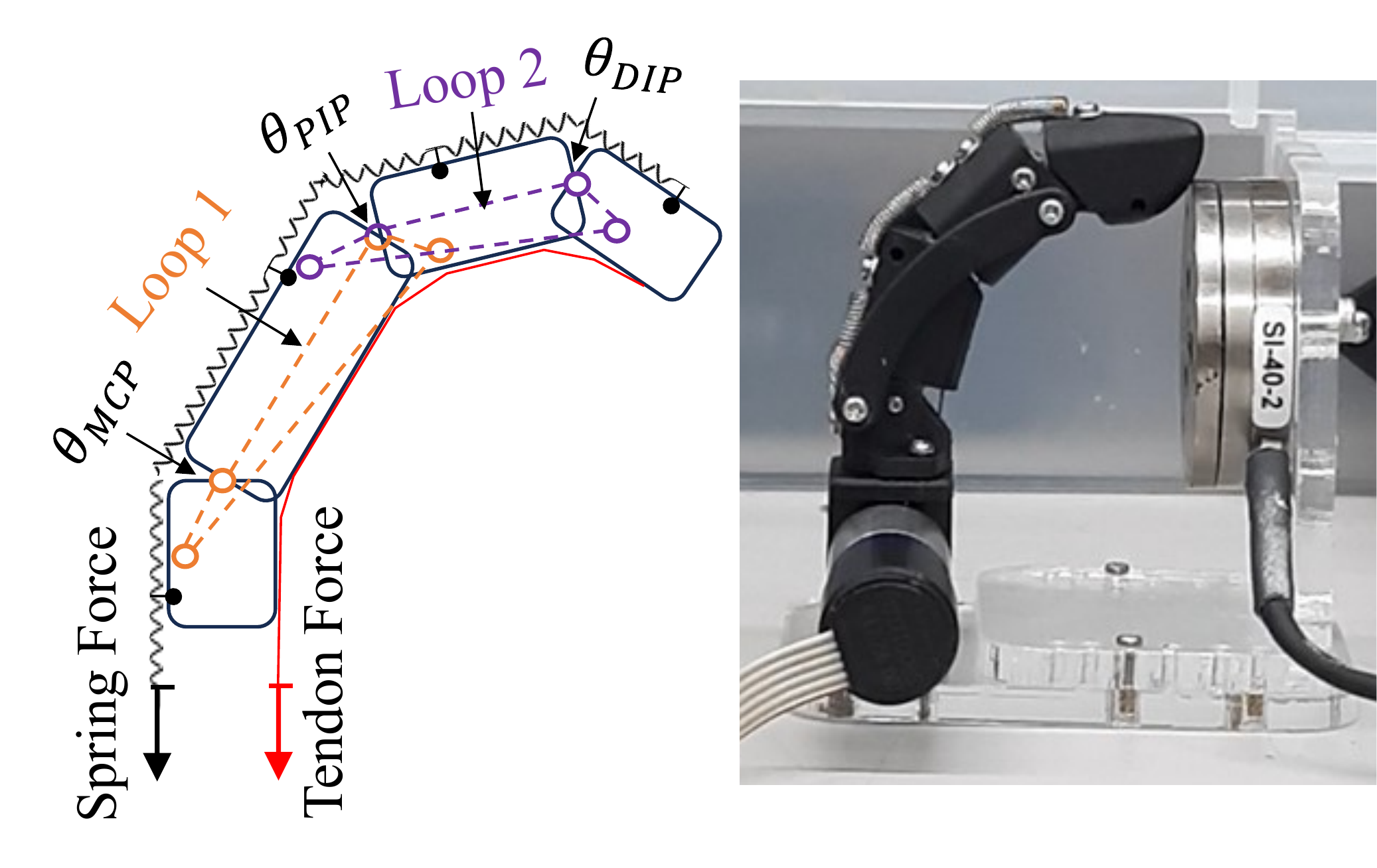}}} \hspace{1cm}
\subfloat[Traditional underactuated tendon driven finger]{%
\resizebox*{5.2cm}{!}{\includegraphics{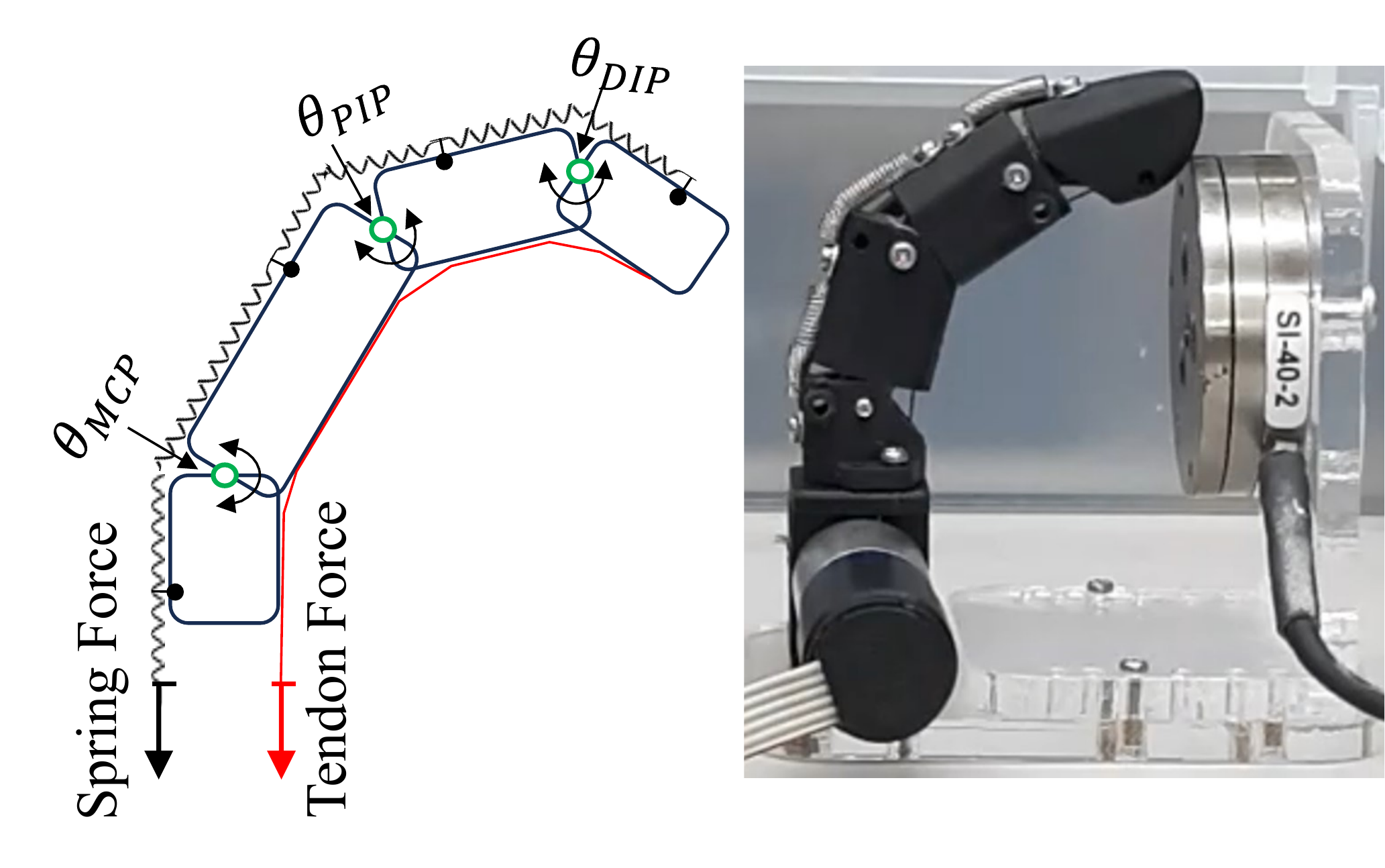}}}
\end{minipage}
\centering
\caption{Comparison between proposed gripper vs existing tendon driven gripper} \label{fig7}
\end{figure}

\subsection{Comparative Analysis with Existing Tendon-Driven Gripper}
\begin{figure}[b!]
\centering
\subfloat[Existing tendon driven finger]{%
\resizebox*{8cm}{!}{\includegraphics{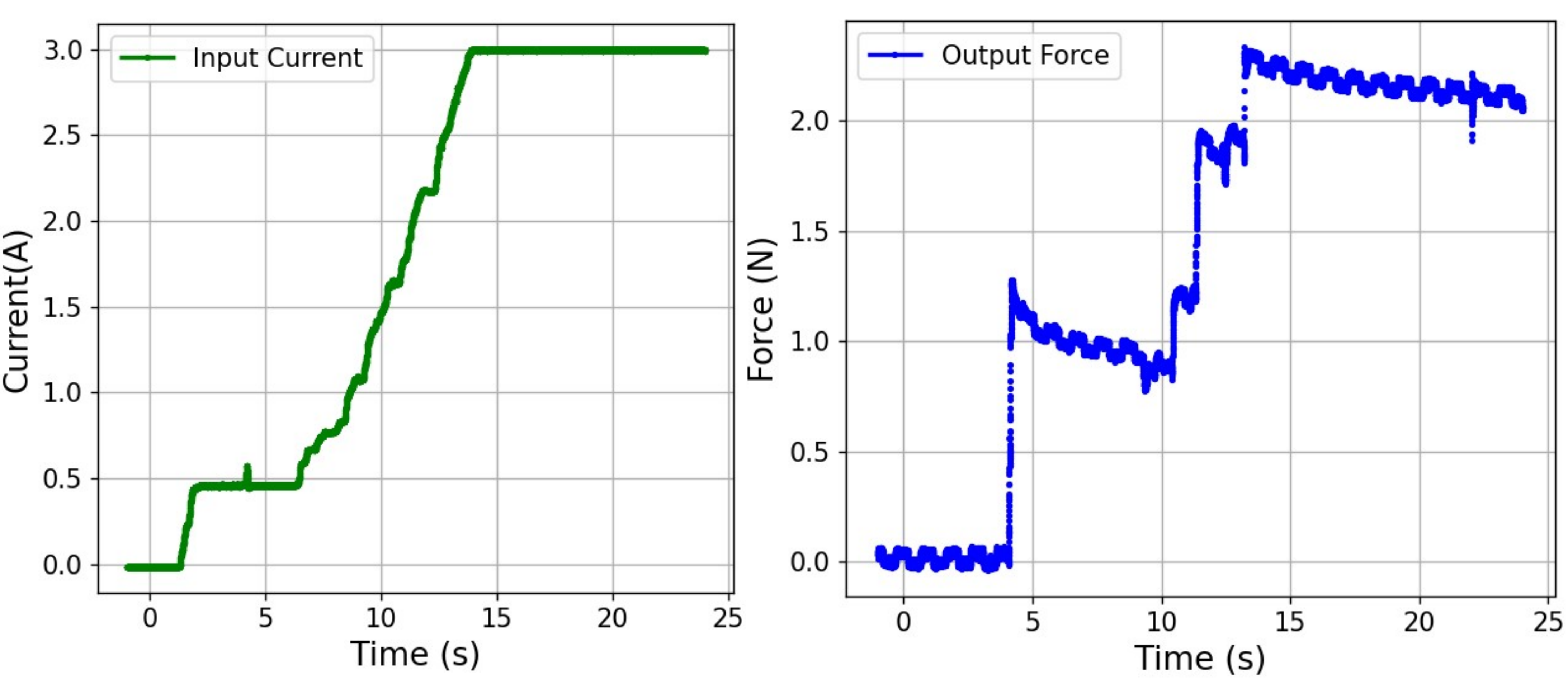}}}
\subfloat[Single tendon driven]{%
\resizebox*{4.2cm}{!}{\includegraphics{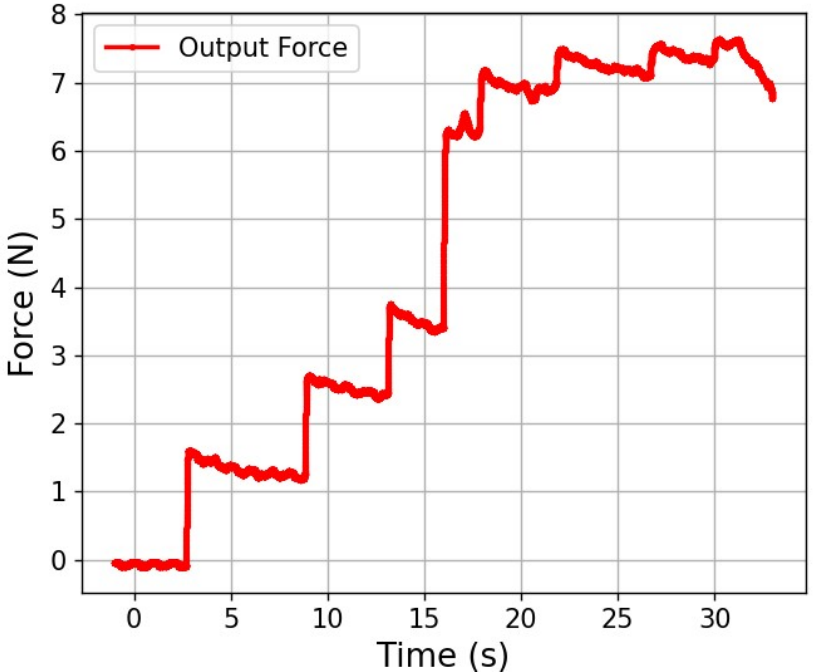}}} \hspace{2cm}
\subfloat[Proposed linkage finger]{%
\resizebox*{8cm}{!}{\includegraphics{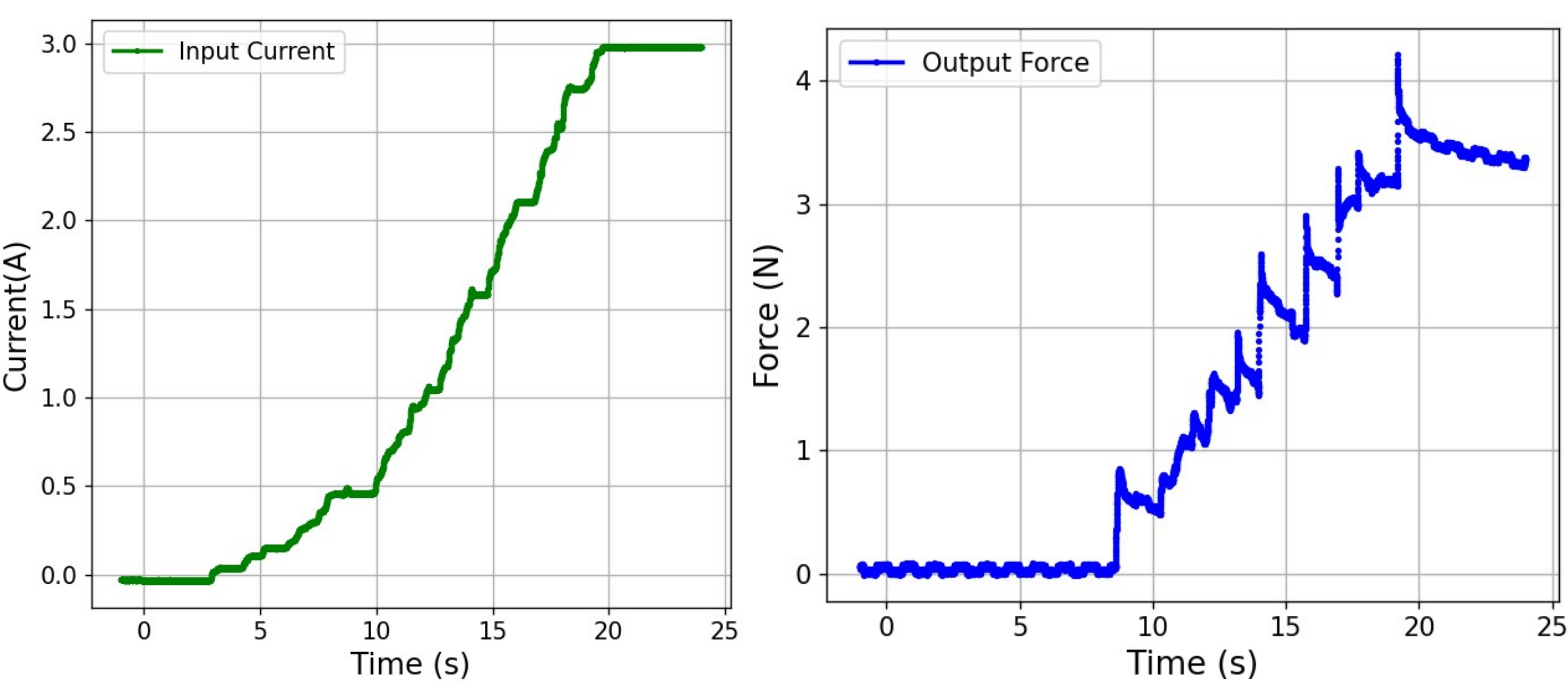}}}
\subfloat[Double tendon driven]{%
\resizebox*{4.2cm}{!}{\includegraphics{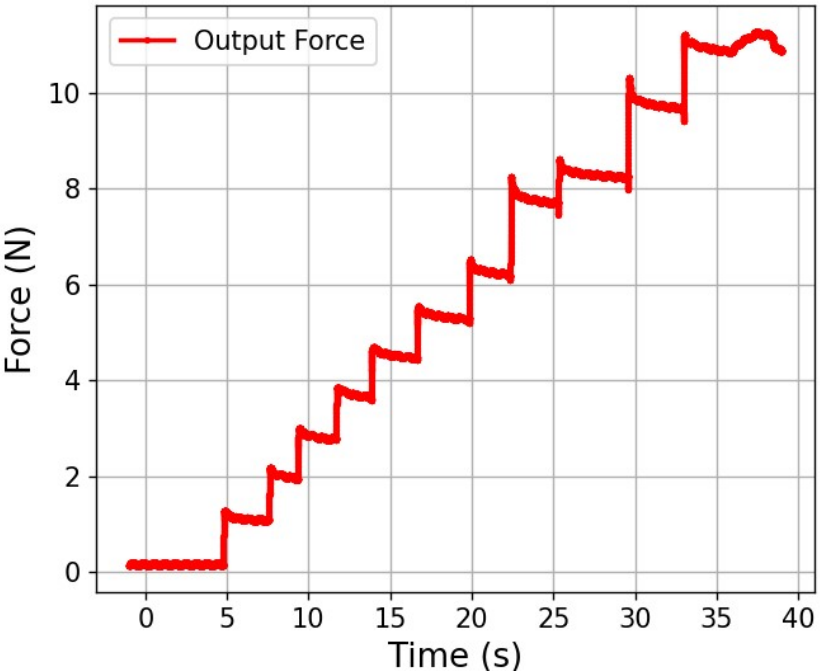}}}
\caption{Experimental results} \label{fig8}
\end{figure}

An experiment was conducted to compare the fingertip force produced by the proposed finger design with a widely used tendon-driven finger design, such as the one utilized in the well-known humanoid robot Nao \cite{aldebaran2023}. The experimental setup, depicted in Figure \ref{fig7}, involved mounting a scaled-down prototype of the anthropomorphic finger on a rigid support and a force sensor to measure output force. The findings, as shown in Figure \ref{fig8} (a) and (c), revealed a 48\% increase in tip force for the proposed finger when supplied with the same power input.

\vspace{-1em}

\subsection{Evaluation of Pinch Force}
Subsequent experiments were carried out to assess the fingertip forces of both single and double-tendon-driven fingers. The experiment involved assessing the tip force at a 90-degree angle of force transmission via tendon tension. After repeating the experiment five times, the average pinch force was determined to be 7.8 N for the single tendon-driven finger and 11.1 N for the double tendon-driven finger (Figure \ref{fig8} (b) and (d)). These results suggest that the gripper, which uses a double tendon actuation for both finger closure and opening, generates a stronger pinch force than the gripper that employs an extension spring for opening. This discrepancy can be attributed to the fact that in the single tendon-driven finger, the tendon must counteract the spring force of the extension spring to close.

\subsection{Evaluation of Gripper Grasping}
To assess the grasping capabilities of the proposed gripper, a variety of tests were carried out. The gripper was tested on a range of objects, including cylindrical items like beverage cans and flat objects such as sheets of paper or cloth (refer to supplementary video). It was also tested on irregularly shaped objects to evaluate its ability to adapt to the surface of the object.

The gripper demonstrated its ability to firmly grasp and hold flat objects with a strong pinch grasp. It was capable of grasping cylindrical objects with diameters ranging from 30 mm to 145 mm, maintaining a secure hold throughout the testing process. The experimental results underscored the effectiveness of the proposed gripper design in achieving both precise pinch and robust cylindrical grasping capabilities.

The gripper was able to generate a substantial pinch force of up to 11.8 N, which allowed it to firmly grasp flat or small objects using a pinch grasp. The results indicated that the proposed gripper design strikes a good balance between precision and strength in its grasping capabilities.

\section{Application of Proposed Gripper for Trouser Dressing-Undressing Assistance}

The feasibility of robot-assisted trouser dressing has been enhanced by two preceding studies, \cite{yamazaki}, \cite{hagiwara}. However, when considering comprehensive support for trouser dressing and undressing, especially for hemiplegic patients in the toilet, the following challenges emerge from these studies:

\begin{itemize}
    \item Lack of practicality when considering support in the confined space of a toilet
    \item Insufficient support for high-need hemiplegic patients
    \item Difficulty in safely grasping trousers worn by humans and providing both dressing-undressing support
\end{itemize}

Introducing a robot system into a typical residential toilet presents significant challenges due to space constraints. Typical household toilets are around 800mm in width, and considering the average shoulder width of a care recipient is about 460mm, this leaves only about 170mm of space on each side for a robot assistive system. The robot system by Hagiwara et al.\cite{hagiwara}, with a hardware body width of about 900mm, is impractical for use in a typical residential building toilet. Thus, conventional research indicates low practicality for robot support in confined toilet spaces.

In addition, almost all the studies shown in Table \ref{table_1} dressing assistance start with pre-grasped non-worn clothes. Most of these studies use conventional rigid two-finger grippers which lack the compliance required for the safe pHRI and difficulty associated with grasping already worn clothing. This includes not only robot assistance for T-shirts and jackets but also trousers dressing assistance. In the case of trouser dressing assistance, the robot gripper already holds the trousers in an ideal shape at the start of the assistance. Hagihara et al.'s study \cite{hagiwara} attempted to grip and move human-worn trousers, but the gripper grasping force was insufficient, causing the trousers to slip from the robot’s grip. 

\begin{figure}[t]
\centering
\subfloat[On the healthy side]{%
\resizebox*{6.5cm}{!}{\includegraphics{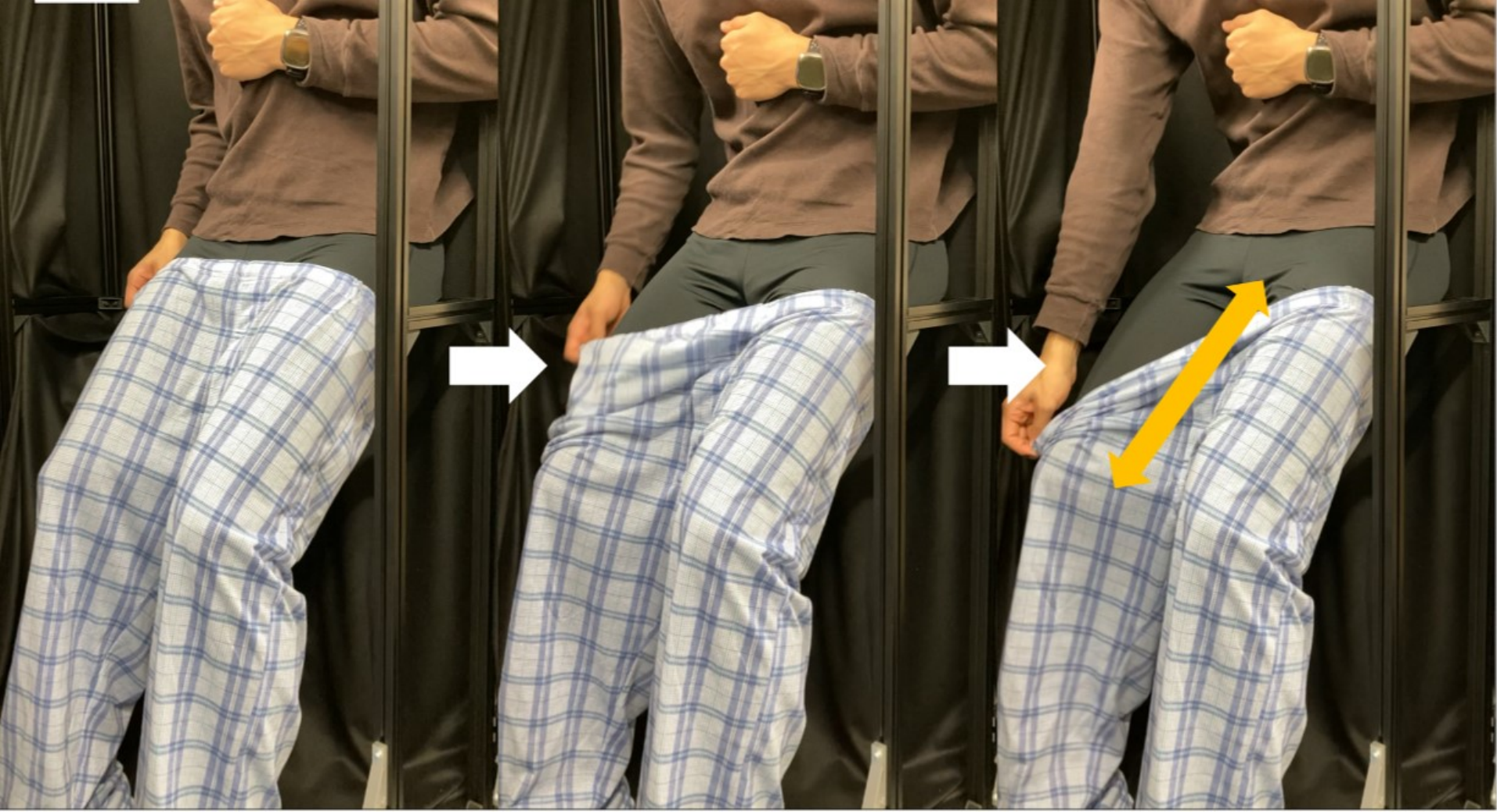}}}\hspace{0.5cm}
\subfloat[On the paralyzed side]{%
\resizebox*{6.5cm}{!}{\includegraphics{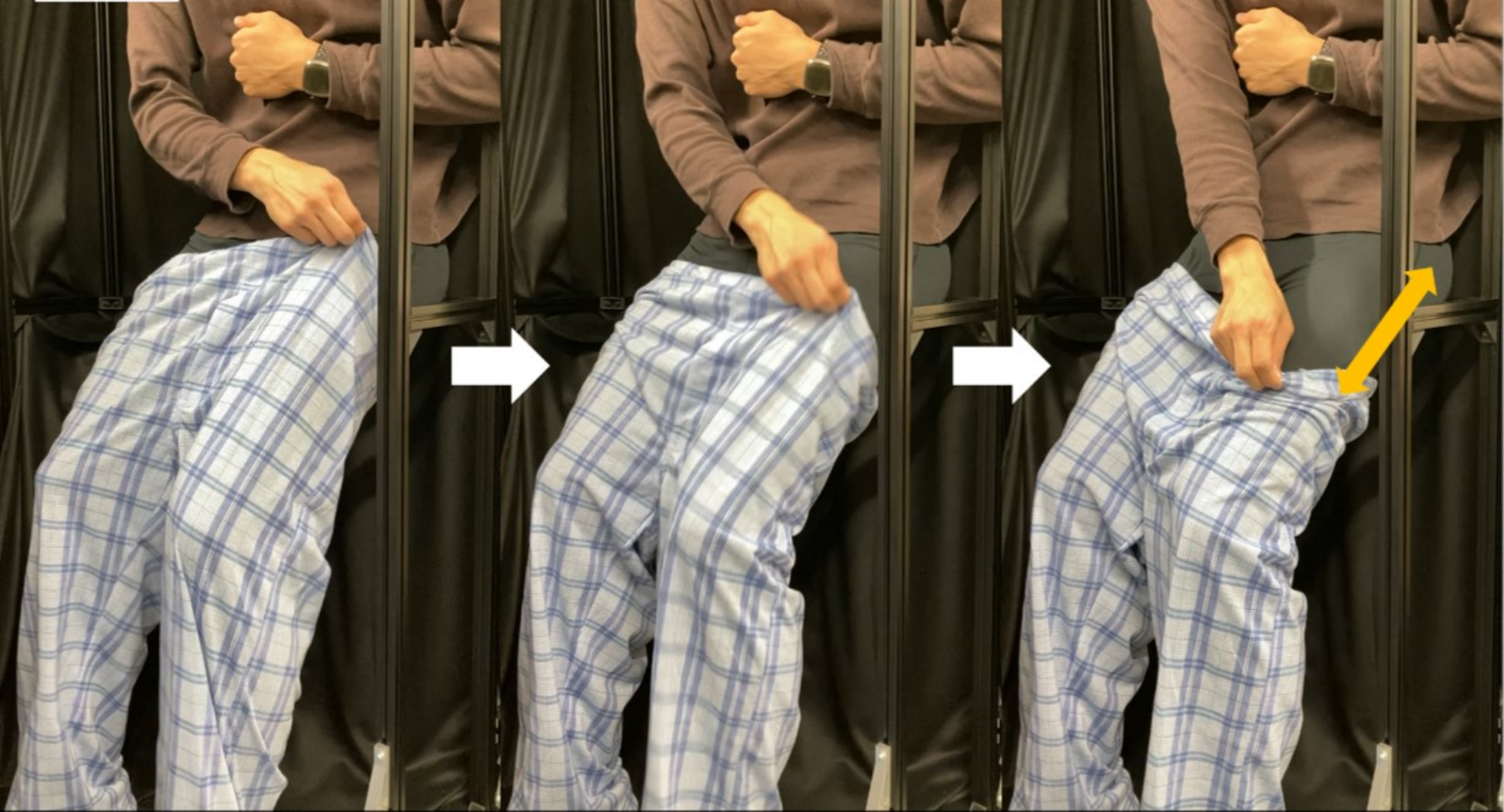}}}
\caption{Extent to which a person with hemiplegia on the left side
of the body can dress and undress their trousers} \label{fig9}
\end{figure}

In the following sections, we detail the application of the proposed gripper and the methodology employed to ensure successful and safe assistance during the dressing and undressing tasks for a hemiplegic individual with minor control over the weakened side and mild cognitive impairment, as represented in Figure \ref{fig9}, during toilet use. The trousers used in this study are cotton pajamas with elastic bands, representative of typical residential attire.

\begin{table}[b]
\tbl{Specifications of proposed gripper for dressing-undressing assistance}
{\begin{tabular}{lc}\toprule
\textbf{Specifications} & \textbf{Proposed Gripper}  \\ \midrule
Weight & 235 g \\ 
Dimensions & $60 mm \times 71 mm \times 177 mm$ $ (W \times L \times H)$ \\ 
Type of Actuation & Single tendon driven \\ 
Actuator & Dynamixel 2XL430-W250 \\ 
DoF & 2 \\ 
Motor Torque & 1.5 Nm (at 12 V, 1.4 A) \\ 
Max No-load Velocity & 12.78 rad/s  \\ 
Fingertip Force & 7.8N  \\ 
Microcontroller &  Dynamixel OpenCR 1.0\\ 
Compliance &  Embodied\\ 
Working space & Figure \ref{fig2} (b)  \\ \bottomrule
\end{tabular}}
\label{table_3}
\end{table}

 \subsection{Integration of proposed gripper with manipulator}

To successfully and safely complete the task of trouser dressing and undressing, we define the following criteria that the gripper must meet:
\begin{enumerate}
    \item The gripper, which comes into physical contact with a person, must be intrinsically safe.
    \item The gripper must be capable of maintaining a grip on the trousers throughout the dressing process.
\end{enumerate}

The proposed single tendon-driven gripper, detailed in Table \ref{table_3}, meets these criteria with its unique design and compliance features tailored specifically for safe robot-assisted trouser dressing and undressing. This gripper demonstrates embodied compliance and responsiveness to external forces, fulfilling the requirement for safe physical human-robot interaction (pHRI). According to the ISO/TS 15066 standard for Robots and Robotic Devices - Collaborative Robots \cite {iso}, the maximum permissible force for thighs and knees should not exceed 220 N for quasi-static contact between human and robotic systems. This threshold is significantly higher than the maximum force that our proposed system can exert, ensuring safety in its application. Additionally, the gripper provides a pinching force of 7.8 N, which is sufficient to securely grasp cotton pajamas and prevent slippage.

\begin{figure}[t]
\centering
\subfloat[Complete assistive system \cite{sugiura}-\cite{inden}]{%
\resizebox*{5.2cm}{!}{\includegraphics{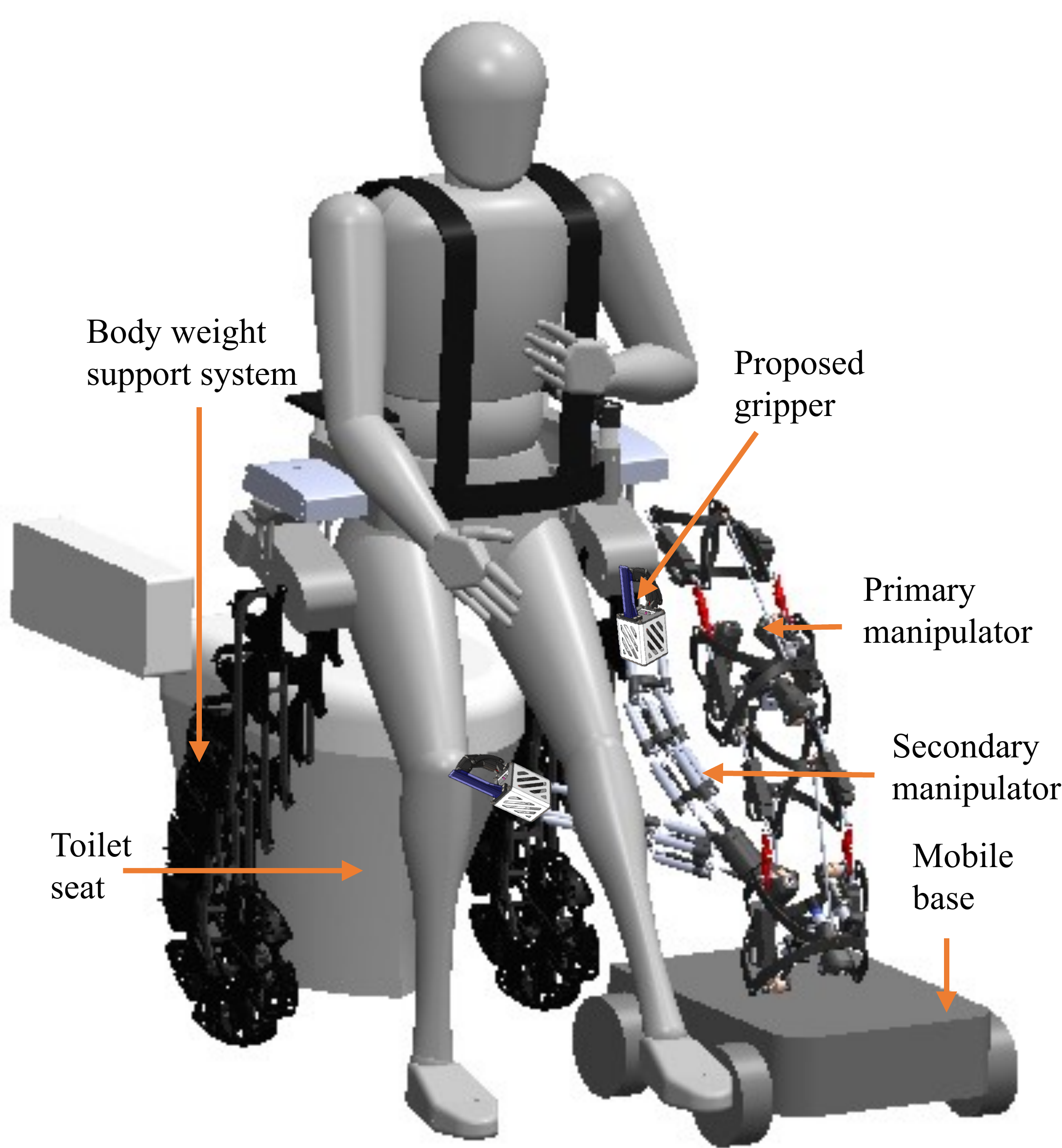}}}
\subfloat[Primary Manipulator]{%
\resizebox*{5.7cm}{!}{\includegraphics{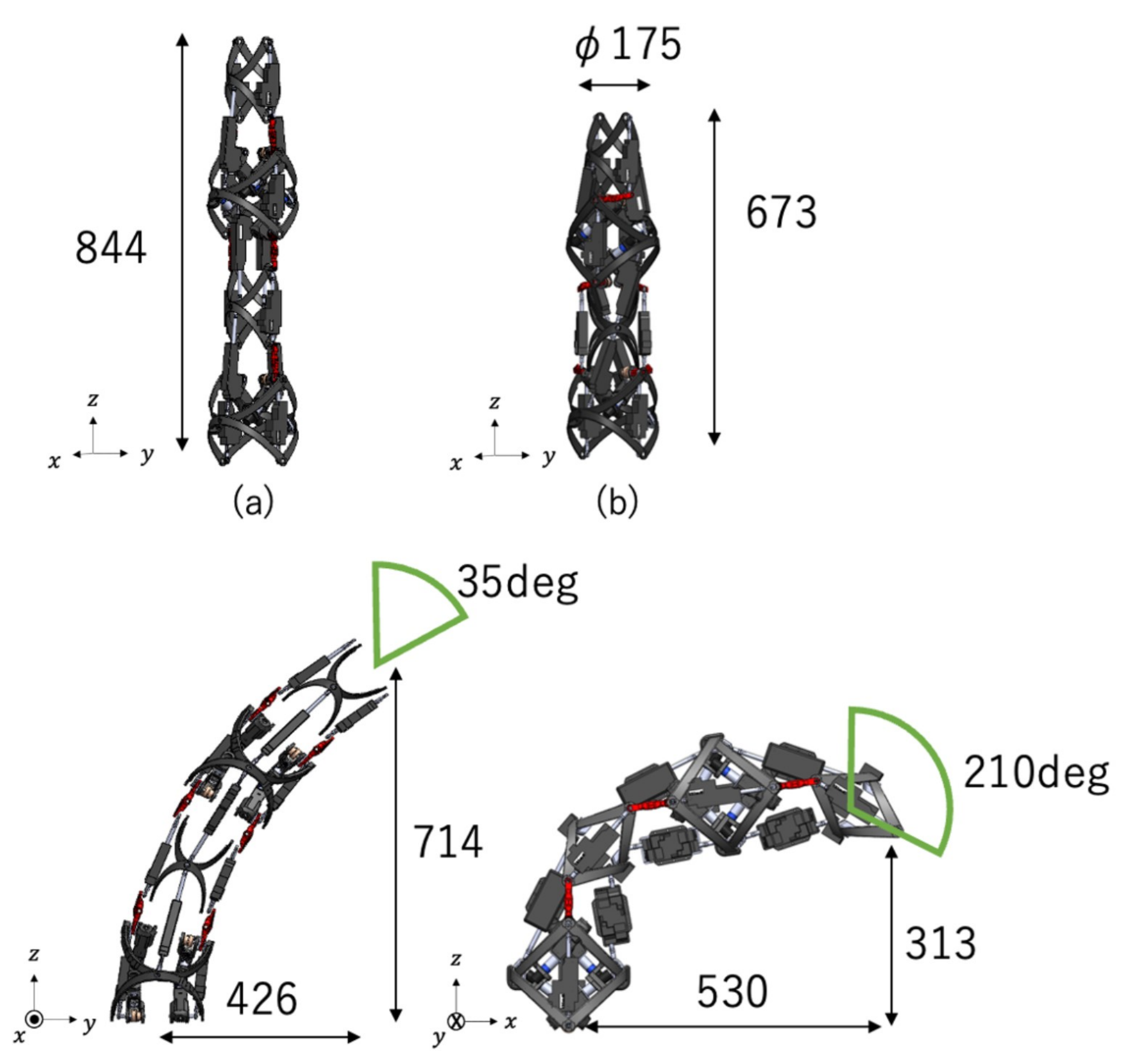}}} 
\subfloat[Secondary Manipulator]{%
\resizebox*{3.5cm}{!}{\includegraphics{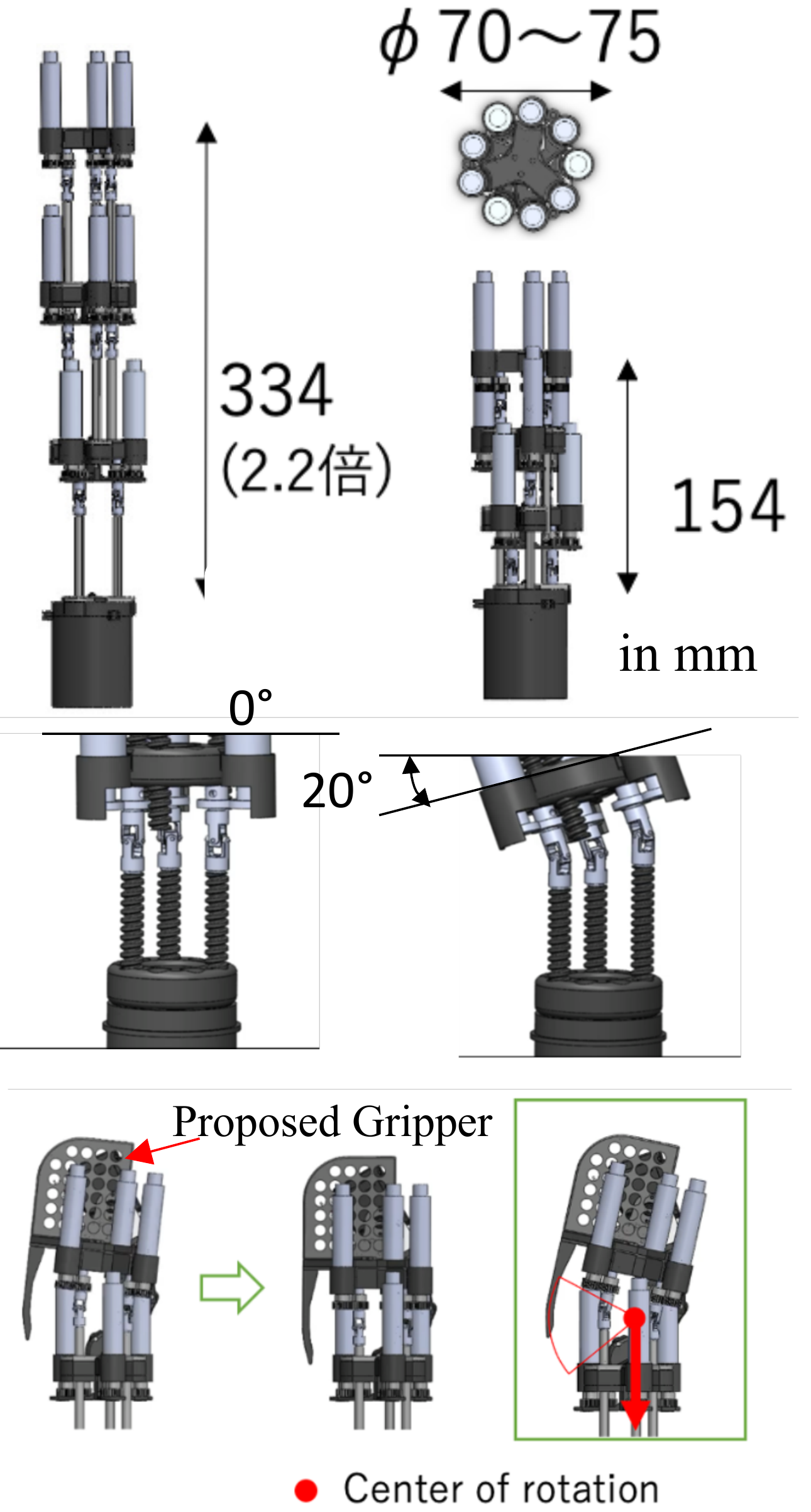}}}
\caption{Comprehensive assistive system developed as part of Moonshot Project} \label{fig10}
\end{figure}

We propose highly adaptable multi-degree-of-freedom robotic arms, specifically designed to aid in the dressing and undressing of trousers in confined spaces. These robots are part of a comprehensive assistive system shown in Figure \ref{fig10} (a) for care homes and welfare facilities, developed as part of Japan’s ‘Moonshot Research \& Development Project’ \cite{moonshot}. This robotic system comprises two types of robot arms, referred to as the primary and secondary manipulator. Both manipulators are equipped with a linkage mechanism that can bend, extend, and retract.

The primary manipulator depicted in Figure \ref{fig10} (b), a multi-degree-of-freedom robotic arm, is 844mm long and 175mm in outer diameter \cite{wakayama}, \cite{yuki}. It has 10 degrees of freedom and a large range of motion due to its ability to bend, extend, and twist. This manipulator is mounted on a mobile omnidirectional robot platform shown in Figure \ref{fig10} (a). The secondary manipulator is mounted on this primary manipulator, which is utilized to position the secondary manipulator accurately for tasks such as dressing and undressing trousers. Due to its hyper-redundant nature, it can position the secondary manipulator at a desired pose, avoiding potential obstacles. In addition, the primary manipulator holds a stereo camera (Intel, RealSense) at the top for visual perception information.

\begin{table}[b]
\tbl{Specifications of manipulators custom-built for dressing-undressing assistance}
{\begin{tabular}{lcc}\toprule
\textbf{Specifications} & \textbf{Primary Manipulator} & \textbf{Secondary Manipulator} \\ \midrule
Weight (with Gripper) & 4.0 kg & 1.5kg \\ 
Max length extension and contraction & 171mm & 180mm \\ 
Max speed extension and contraction & 0.038m/s & 0.09m/s \\ 
Speed while grasping & - & 0.06m/s \\ 
Tip Force & 3.0kg & 1.5kg \\ 
Tip Rigidity (Max extended length) & 6.67mm/kg & 10mm/kg  \\ 
Accuracy of tip position & ± 20mm & - \\ 
Working space & Figure \ref{fig11} (b) & Figure \ref{fig11} (c) \\ \bottomrule
\end{tabular}}
\label{table_4}
\end{table}

The secondary manipulator is a smaller robot arm with a total length of 334mm and an outer diameter of 75mm. It has a total of 9 degrees of freedom for extension, contraction, and bending, realized by a 3-stage parallel link lead screw mechanism \cite{inden}.  As illustrated in Figure \ref{fig10} (c), the arm can be extended, retracted, and bent by adjusting the length of the three lead screws in each stage. It can extend and contract by 180 mm, which is sufficient for the required movement of trousers during toilet usage (about 160 mm). This manipulator is responsible for providing actual trouser dressing and undressing assistance to the left side of a hemiplegic individual, as shown in Figure \ref{fig11} (a). The proposed gripper is mounted on this manipulator to provide this assistance. This enables the system to assist in dressing and undressing by grasping and moving clothing in close proximity to the human body, even in narrow spaces of about 170mm in width.

\section{Experimental Evaluation of Trouser Dressing-Undressing Assistance}

\begin{figure}[t]
\centering
\subfloat[Assistive system in toilet environment]{%
\resizebox*{8cm}{!}{\includegraphics{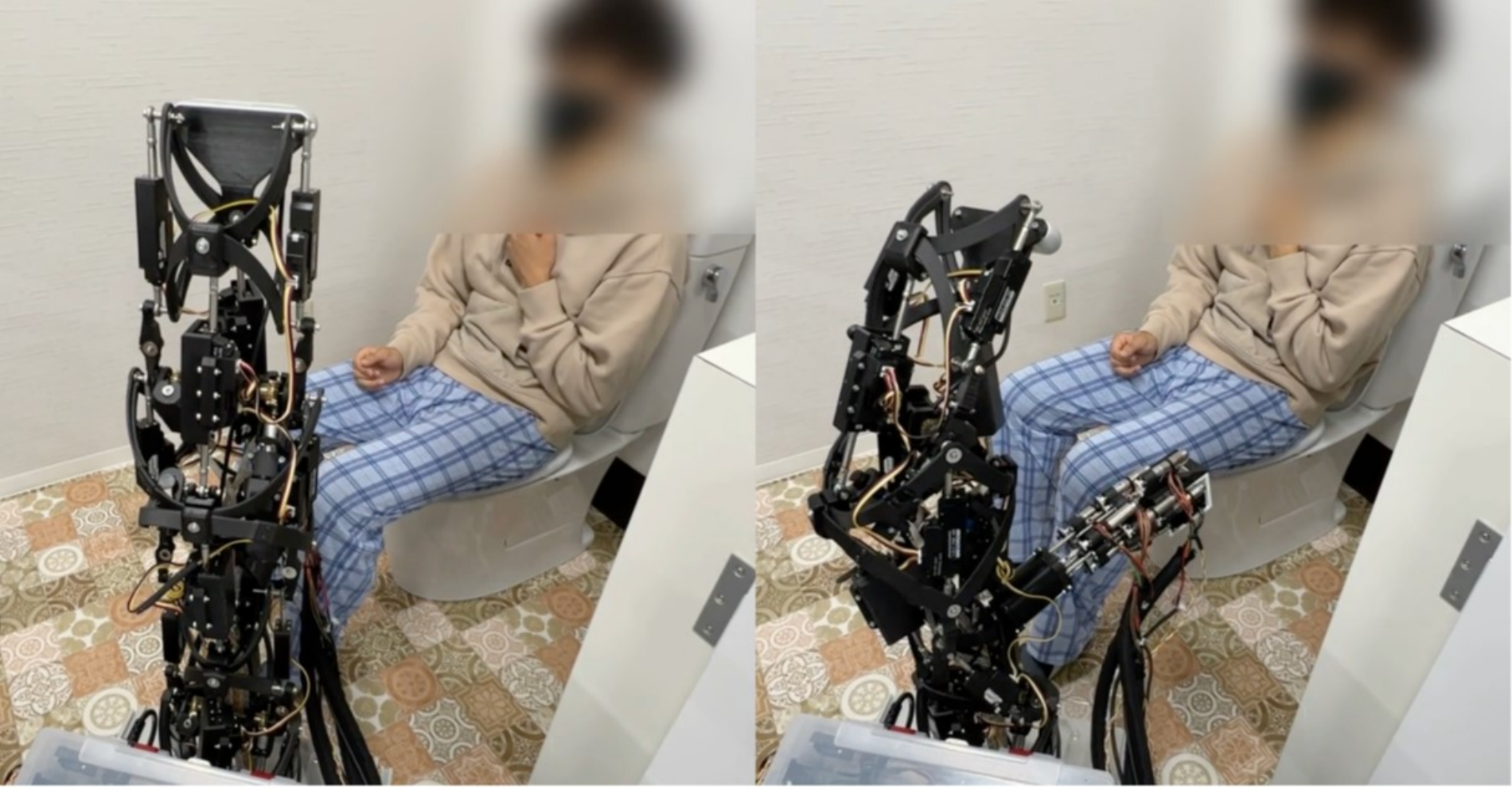}}}\hspace{1cm}
\subfloat[Primary manipulator workspace]{%
\resizebox*{8cm}{!}{\includegraphics{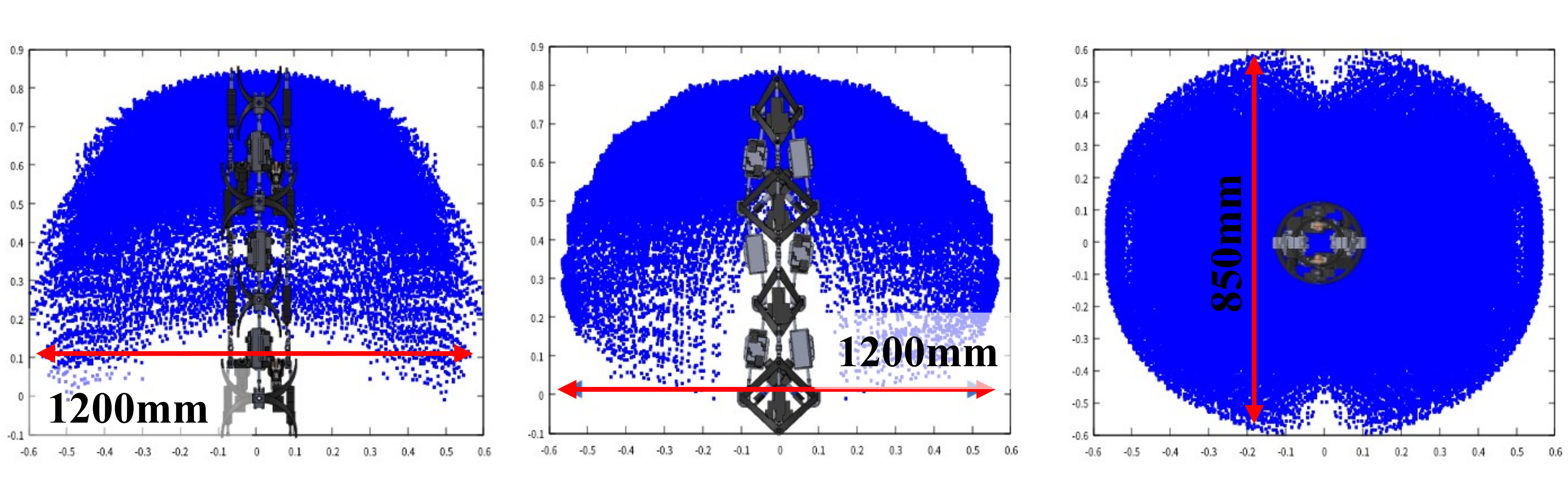}}} \hspace{1cm}
\subfloat[Secondary manipulator workspace]{%
\resizebox*{8cm}{!}{\includegraphics{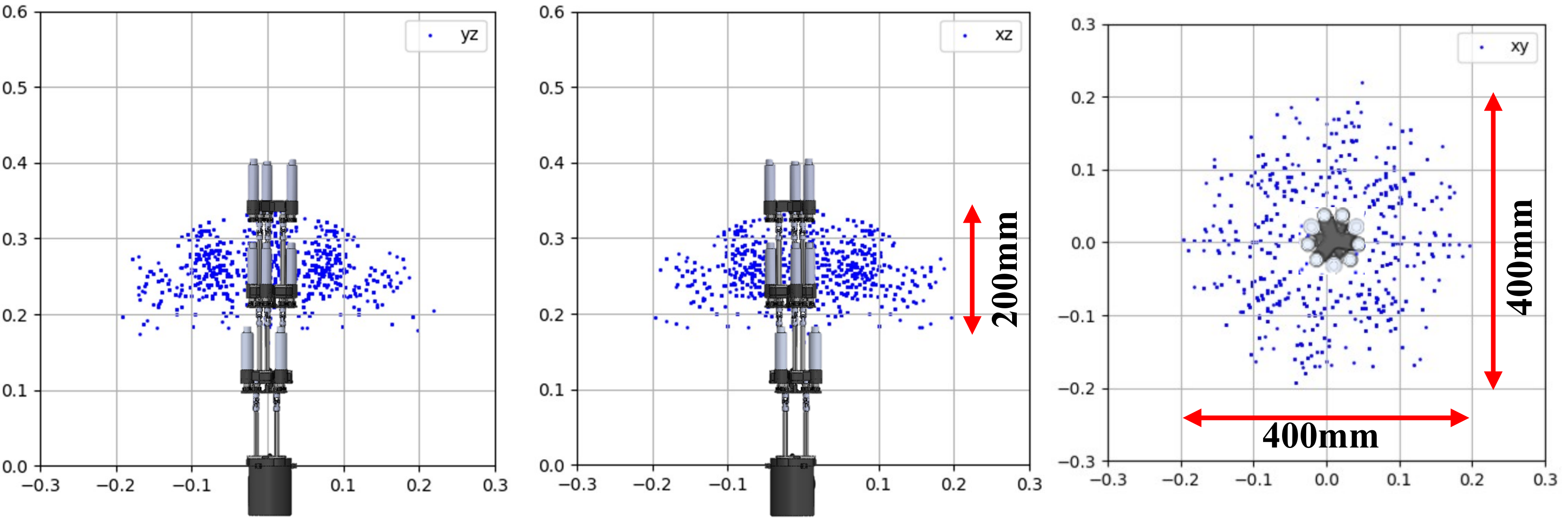}}}
\caption{Reaching to trouser in narrow toilet space} \label{fig11}
\end{figure}

In this section, we aim to verify the ability of the proposed system to aid in dressing-undressing trousers in the restroom. The experimental environment shown in \ref{fig11} (a) replicates a typical home environment for testing rehabilitation support and demonstrating the functionality of care robots. The specification of the verification environment is as follows: The toilet measures 1.5m in width and 1.85m in length. The distance from the toilet to the wall is 0.92m at the entrance, 0.51m on the right side (assuming a left-side paralysis), and 0.61m on the left side.

The verification process was conducted in the following sequence:
\begin{enumerate}
    \item The secondary manipulator, with the assistance of the primary manipulator, positions to the appropriate location, utilizing its extension and retraction features.
    \item The secondary manipulator then assists in dressing and undressing the trousers.
\end{enumerate}
Here, it is assumed that the primary manipulator is already positioned at the desired location of the toilet by the mobile platform to the paralyzed side of the hemiplegic care recipient (refer to supplementary video).

\subsection{Positioning Secondary Manipulator for Assistance}
 The first step in the verification sequence details the method in which the primary manipulator maneuvers the secondary manipulator into the desired pose to provide effective assistance. As depicted in \ref{fig11} (a), the primary manipulator employs its hyper-redundant DoF to position the secondary manipulator in a confined space.

The secondary manipulator, placed on the first primary manipulator, was successfully positioned for support near the knee of the hemiplegic individual. In the experimental environment, where the toilet and the wall on the paralyzed side were wide (approximately 500 mm), the primary manipulator was moved near the front of the paralyzed side of an individual using the mobile platform. From this position, the secondary manipulator was maneuvered near the front of the patient's knee, along the thigh of the paralyzed side, in a suitable position for support. This method of support, where the primary manipulator is moved close to the patient's knee and the secondary manipulator is positioned along the thigh on the paralyzed side, was found to be feasible even in a narrow space, such as a typical household toilet. This is evident in Figure \ref{fig11} (a), where limited space is visible between the right-hand white wall and the patient seated on the toilet.

\begin{figure}[t]
\centering
\includegraphics[scale=0.175]{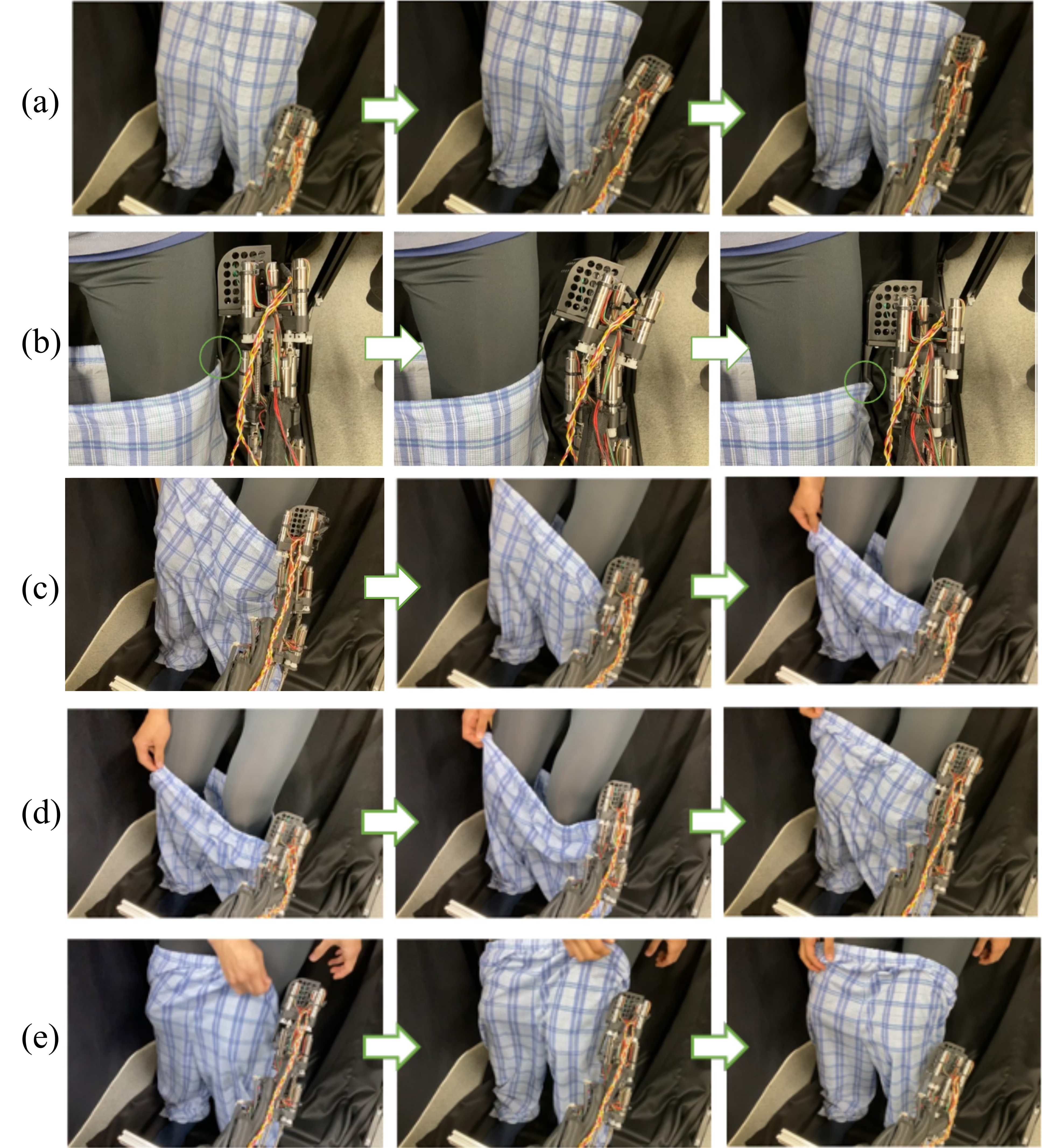}
\caption{Trouser dressing-undressing assistance} \label{fig12}
\end{figure}

\subsection{Trouser dressing-undressing by secondary manipulator with proposed gripper}
Next, we detail the process of the secondary manipulator with the proposed gripper for grasping and moving trousers. The sequence of support actions is as follows:
\begin{enumerate}
    \item  Starting from its most compact state (initial state), the secondary manipulator extends towards the waistband of the trousers, utilizing its multiple degrees of freedom.
    \item Leveraging its degrees of freedom, the gripper's fixed thumb is carefully inserted between the waistband and the human body, and the trousers are grasped.
    \item The trousers, once gripped, are lowered to the care recipient's knee.
    \item The trousers are then pulled back up to the point where they were initially grasped.
    \item The gripper's inserted fixed thumb is removed carefully.
    \item Finally, the manipulator returns to its initial position.
\end{enumerate}

Figure \ref{fig12} shows the trouser dressing undressing support provided by the secondary manipulator with gripper, by grasping, lowering, and then raising the trouser by 170mm (refer to supplementary video). Where the entire system was teleoperated using a joystick and vision information from the stereo camera mounted on the primary manipulator. During the validation, the subjects wore spats. Examining each support action individually, starting with (a), from the initial position, the secondary manipulator's 180mm extension was enough to reach the trouser's waistband at the mid-thigh level. This is within the range where the subject's healthy arm is able to effectively lower the trousers, as depicted in Figure \ref{fig9} (b). Next, in (b), we can see that the gripper fixed thumb was inserted in the trouser waistband by utilizing the rotational degree of freedom shown in Figure \ref{fig10} (c). 

Upon visual confirmation of the fixed thumb's insertion, the gripper employs its compliant active finger to firmly grasp the trouser waistband with a pinch force of about 10N. After the trousers are grasped, as shown in Figure \ref{fig12} (c), the secondary manipulator lowers the trousers by about 170mm on the paralyzed side along with the care recipient lowering the right side. Similarly, in step (d), the manipulator raises the trouser to the starting position. Finally, in step (e), at the end of the support, the manipulator successfully returned to the initial position in such a way as to avoid contact with the human body.

\subsection{Experimental result and comparison}
 Table \ref{table_5} illustrates the comparison and evaluation of trouser dressing-undressing assistance provided by the proposed system, prior study, and that given by human caregivers. The participants in this study were four healthy young men. All participants provided their consent prior to the commencement of the experiments. The study was also approved by the Ethics Committee of Nagoya University (Approval No. 22-16). While the proposed system is designed to assist individuals with hemiplegia, this paper verifies the basic assistive effectiveness of the proposed system through experiments involving healthy subjects. To simulate the condition of a left-hand hemiplegic patient, the subjects were instructed to use only their right hand during the experiment. The methodology detailed in Section 6.2 was employed for the dressing and undressing assistance.

 \begin{table}[t]
\centering
\tbl{Comparison and evaluation of trouser dressing-undressing assistance}
{\begin{tabular}{P{2.19cm}M{0.8cm}M{1cm}M{0.75cm}M{1.25cm}M{0.8cm}M{1cm}M{0.8cm}M{1.1cm}}
\toprule
Assistance & \multicolumn{4}{c}{Dressing} & \multicolumn{4}{c}{Undressing} \\ \midrule
 & No. of trials & No. of success & Success rate & time\textsuperscript{a}($s$)& No. of trials & No. of success & Success rate & time\textsuperscript{a}($s$)\\ \cmidrule{2-9}

Prior study \cite{hagiwara} &10&9&90\%&unlisted&7&0&0\%& unlisted\\
Proposed system&4&4&100\%&3.0&4&4&100\%&3.1\\
Human &4&4&100\%&2.0&4&4&100\%&2.0\\
\bottomrule
\end{tabular}}
\flushleft
\tabnote{\textsuperscript{a} Average task completion time in seconds.}
\label{table_5}
\end{table}

In all four trials, the proposed gripper successfully grasped and handled cotton pajamas without slippage. Moreover, the subjects expressed that they did not feel any pain, discomfort, or fear during the experiment, showcasing the ability of the proposed gripper to perform the safe pHRI. In the prior research \cite{hagiwara}, the robot’s ability to assist with trouser dressing was limited due to constraints in degrees of freedom, output force, and grasping force. Conversely, the newly proposed system has demonstrated a high success rate in assisting with both dressing and undressing trousers in confined spaces. However, the time taken by the proposed system for assistance is still longer than that of a human caregiver. The limiting factor here is the extension-retraction speed of the secondary manipulator. Moving forward, we aim to fully automate the process by incorporating additional sensors and vision perception, alongside implementing compliant control to ensure the highest level of safety during human-robot interaction. Our ultimate goal remains the development of a fully autonomous dressing-undressing assistance system that prioritizes user safety above all else.

\section{Limitations}
Despite the high success rate of the proposed robotic system in assisting with dressing and undressing, it’s important to recognize the study’s limitations.

The gripper’s fingers are compliant, but the fixed thumb, even though it’s 3D-printed from a relatively elastic TPU material, remains somewhat rigid. Future iterations could benefit from replacing the fixed thumb with a more compliant mechanism, eliminating the need for trouser insertion. Alternatively, implementing a compliant impedance control on the secondary manipulator could further enhance the safety of this physical human-robot interaction.

Currently, the system requires teleoperation and visual monitoring for the gripper's position and orientation control. Our future goal is to automate this assistance system. For example, we could use Simultaneous Localization and Mapping (SLAM) to enable the mobile platform to autonomously navigate indoor environments and reach care recipients in the toilet. Additionally, the stereo camera mounted on the primary manipulator could be used to estimate the human body’s posture and plan the robot arm’s motion path autonomously. Moreover, the perception information from the camera could be used to find the ideal grasp point and verify the successful grasping of trousers. Furthermore, the subjects of the study could be increased with broad demographics. Finally, the different types of trousers and fabric materials could be tested. 

Despite these limitations, this work demonstrates the feasibility of the proposed approach and offers a comprehensive solution for assisting the elderly or individuals with disabilities in confined spaces, a task not previously achieved.

\section{Conclusion}
Our research presents a novel gripper system, specifically designed for robot-assisted dressing and undressing of trousers, catering to the needs of hemiplegic individuals. The gripper's features, including embodied compliance, controllability, and adequate grasping force, collectively tackle the challenges in trouser manipulation, ensuring safe human-robot interaction.

The proposed gripper was integrated into a robotic manipulator system and paired with a mobile platform. This combination provides a comprehensive solution to assist hemiplegic individuals in their dressing and undressing tasks. The gripper's compact and lightweight design enhances its usability, making it easy to incorporate into existing care environments. Our experimental evaluation demonstrates the gripper's effectiveness and reliability in performing trouser manipulation tasks in confined spaces.

In conclusion, the successful development and evaluation of the gripper system mark significant progress in assistive robotics for the elderly and individuals facing difficulties in their activities of daily living. The proposed assistive system fills the gaps with preceding studies and aims to foster independence and self-efficacy among individuals with hemiplegia. We hope that this will contribute toward an improved quality of life and reduce the burden on caregivers. Future work will focus on refining the gripper's design, conducting extensive user trials, and exploring its integration into real-world care settings.

 \section*{Funding}
This work was supported by JST [MoonshotR\&D] [Grant Number JPMJMS2034].

\section*{Notes on contributor}
\subsection*{Jayant Unde}
\textbf{Jayant Unde} completed his B.Tech in Mechanical Engineering from COEP Technological University, Pune, India, in 2018. He then went on to earn his M.E. from Nagoya University in 2023. Currently, he is a MEXT scholar and is pursuing his Ph.D. at the Department of Micro-Nano Mechanical Science and Engineering, Nagoya University. His research interests include assistive robotics, mechanism design, grippers for cloth manipulation, and soft robotics. He was the recipient of the Best Paper Award at the IEEE International Conference on Robot and Human Interactive Communication (RO-MAN) in 2023.

\subsection*{Takumi Inden}
\textbf{Takumi Inden} completed his B.E. and M.E. degrees from the Department of Micro-Nano Mechanical Science and Engineering at Nagoya University in 2021 and 2023, respectively. His areas of interest include manipulator design, hyper-redundant mechanisms, and robot control.

\subsection*{Yuki Wakayama}
\textbf{Yuki Wakayama} obtained his B.E. and M.E. degrees from the Department of Micro-Nano Mechanical Science and Engineering at Nagoya University in 2022 and 2024, respectively. His research is focused on manipulator control, grasping mechanisms, and mobile robotics.

\subsection*{Jacinto Colan}
\textbf{Jacinto Colan} received the B.S. degree from the National University of Engineering, Lima, Peru, in 2010, and the M.E. and Ph.D. degrees in micro-nano mechanical science and engineering from Nagoya University, Nagoya, Japan, in 2018 and 2021, respectively. He is currently a Postdoctoral Researcher with Nagoya University. His main research interests include medical robotics, human–robot interfaces, and intelligent assistive systems. He received the Best Paper Award from the IEEE International Symposium on Micro-Nano Mechatronics and Human Science (MHS 2019 and MHS 2022).

\subsection*{Yaonan Zhu}
\textbf{Yaonan Zhu} received the B.S. degree in automation from Northeastern University, Shenyang, China, in 2015, the M.S. degree in computer science from the University of Science and Technology of China, Hefei, China, in 2018, and the Ph.D. degree in robotics from Nagoya University, Nagoya, Japan, in 2021.He is currently a Designated Assistant Professor with Nagoya University. His research interests include telerobotics and teleoperation, haptics and haptic interfaces, human-centered robotics, and wearable robotics.

\subsection*{Tadayoshi Aoyama}
\textbf{Tadayoshi Aoyama} received the B.E. degree in mechanical engineering, the M.E. degree in mechanical science engineering, and the Ph.D. degree in micro-nano systems engineering from Nagoya University, Nagoya, Japan, in 2007, 2009, and 2012, respectively. He was an Assistant Professor with Hiroshima University, Japan, from 2012 to 2017, and at Nagoya University, from 2017 to 2019. He was a PRESTO Researcher at JST, from 2018 to 2022. He is currently an Associate Professor with the Department of Micro-Nano Mechanical Science and Engineering, Nagoya University. His research interests include macro-micro interaction, VR/AR and human interfaces, AI-based assistive technology, micromanipulation, and medical robotics.

\subsection*{Yasuhisa Hasegawa}
\textbf{Yasuhisa Hasegawa} received the B.E., M.E., and Ph.D. degrees in robotics from Nagoya University, Nagoya, Japan, in 1994, 1996, and 2001, respectively. From 1996 to 1998, he was with Mitsubishi Heavy Industries Ltd., Japan. He joined Nagoya University, in 1998. Then, he moved to Gifu University, in 2003. From 2004 to 2014, he attended the University of Tsukuba. Since 2014, he has been with Nagoya University, where he is currently a Professor with the Department of Micro/Nano Mechanical Science and Engineering. His research interests include motion-assistive systems, teleoperation for manipulation, and surgical support robots.

\end{document}